%% file: master-document.tex
\pdfoutput=1
\documentclass{aimsessay}
\usepackage[utf8]{inputenc}
\usepackage[T1]{fontenc}
\usepackage{lmodern}
\usepackage[round]{natbib}
\usepackage{amsmath}
\usepackage{amsfonts}
\usepackage{amssymb}
\usepackage{mathtools}
\usepackage{latexsym}
\usepackage{parskip}
\usepackage{enumerate}
\usepackage{tikz}
\usepackage{lipsum}
\usepackage{hyperref} 
\makeatletter\def\thm@space@setup{%
  \thm@preskip=1.2\parskip \thm@postskip=0pt
  }

\makeatother
\usepackage{caption} 
\usepackage{subcaption} 
\usepackage{booktabs} 
\usepackage{array}    
\usepackage{float}  
\usepackage{moreverb}   
                        
\usepackage{algorithm}
\usepackage[noend]{algpseudocode}
\makeatletter
\def\BState{\State\hskip-\ALG@thistlm}
\makeatother 
\swapnumbers 
\theoremstyle{plain}

\theoremstyle{definition}

\newtheorem{defn}[subsection]{Definition}

\numberwithin{equation}{section}
\title{An empirical investigation into the properties of standard word embeddings}
\author{Salomon Kabongo Kabenamualu (salomon@aims.ac.za)\\
African Institute for Mathematical Sciences (AIMS)\\
\\
{\small Supervised by: Professor Etienne Barnard}\\
{\small North-West University}%
}
\date{{\small 23 May 2019}\\%
  {\scriptsize\it Submitted in partial fulfillment of 
    a structured masters degree at AIMS South Africa}\\%
  \vspace{0.5cm}{\includegraphics{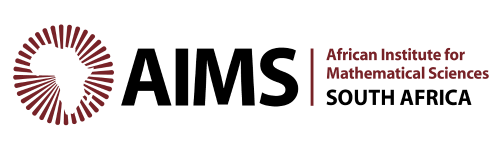}}}
\begin{document}
\pagestyle{empty}
\maketitle
%
\pagenumbering{roman}
\input{abstract}

\tableofcontents
\newpage
%
\pagenumbering{arabic}
\pagestyle{myheadings}
\input{chapter1}			

\input{chapter2} 			
\input{chapter3} 			
\input{chapter4}
\input{chapter5} 			
\input{acknowledgement}
\renewcommand{\bibname}{References}
\nocite{*}
\bibliographystyle{myabbrvnat}
\bibliography{references}
\addcontentsline{toc}{chapter}{References}
\end{document}

%% file: abstract.tex
\chapter*{Abstract} 
\addcontentsline{toc}{chapter}{Abstract}

%
%
%


The embedding of word sequences into continuous vector spaces has been one of the most important developments in Natural Language Processing in the recent past. Such embeddings have found application in areas such as Automatic Speech Recognition, Machine Translation, Sentiment Analysis and many more. This essay reviews the various mechanisms that have been proposed for the calculation of word embeddings, investigates popular toolkits and embedding matrices that are available in the public domain, and experiments with one or more selected implementations to better understand their characteristics.

La representation vectoriel continue de mots a été l'un des développements les plus importants dans le domaine du traitement automatique du langage naturel au cours des dernières années. Ces representations ont trouvé application dans des domaines tels que la reconnaissance vocale, la traduction automatique, l'analyse des sentiments, etc. Ce travail passe en revue les différents mécanismes proposés pour le calcul des ces vecteurs de mots, étudie les kits d'outils populaires et les matrices disponibles publiquement en ligne, et expérimente avec une ou plusieurs implémentations sélectionnées pour mieux comprendre leurs caractéristiques.

\textbf{Keywords}: \textit{NLP, Embeddings, Neural Networks, Machine Learning}.




\vfill
\section*{Declaration}
I, the undersigned, hereby declare that the work contained in this research project is my original work, and that any work done by others or by myself previously has been acknowledged and referenced accordingly.

\includegraphics[height=2cm]{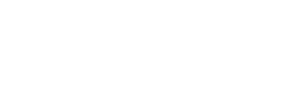} \newline \hrule
Salomon KABONGO KABENAMUALU, 23 May 2019

%% file: chapter1.tex
\chapter{Introduction}\label{chap1}
Human being are equipped with amazing features to understand, analyze and interpret events\textemdash natural intelligence or ability. However, under some circumstances, natural ability can be limited. It is possible to construct artificial machines with tremendous processing resources.  Mimicking such features with artificial intelligence is a great achievement human race can ever realize. This essay reviews machine learning, a subfield of artificial intelligence, in this regards.
Machine learning covers different tasks for various applications. Here we are interested in Natural Language Processing.

Humans learn how to read and write from an early age in a seemingly effortless manner. However making a computer to do the same is not obvious. The real challenge in artificial intelligence has proved to be solving tasks that are easy for people to perform but hard for people to describe formally, problems that we solve intuitively, that feel automatic, like recognizing spoken words or faces in an images \cite{Goodfellow-et-al-2016}. 

In this project we will address one of those intuitive problems. \textit{How can a computer deal with text, and use it effectively ?}.

Early approaches in natural language processing (NLP) were based on human-curated rules, but this lead to several problems, as the number of rules required to effectively capture semantic and syntactic meaning on text quickly exploded. 

As a solution to the challenge caused by the rule-based approach, the \textit{statistical revolution}(early 1990s) marked the moved from rule-based approach to statistical NLP approach---machine learning \citep{Johnson:2009:SRC:1642038.1642041}. Over the last $20$ years, this approach has proven very successful.

Talking of statistics, means the use of numbers, this fact make it easy and natural as the mathematics involved in the construction of a neural networks (as we will see later in this essay) requires multiplication of the input with the weight added to the bias, but in NLP most input data is textual. So the question that arises is \textit{how to numerically represent non numerical value while capturing insight between words ?}

This work will experiment several word representation, and give overviews of various mechanics that have been proposed for their calculations as well as experimentation on a sentiment analysis task.

Several works \cite{Goldberg:2016:PNN:3176748.3176757}, \cite{Bengio03aneural} suggest the use of more advance deep learning architecture(1D-CNN, RNN) to model sequential data, but we will demonstrate how the use of pre-trained word embeddings in combination with the weighted continuous bag of words can produce state-of-the-art or nearly state-of-the-art results. 

Throughout this report the word \textit{feature} is used to refer to a linguistic input (word, part-of-speech, $\ldots$) to our Neural Network. 

Bold upper case letters, will be used to represent matrices ($\mathbf{X},\mathbf{Y}$), and lower-case letters for vectors representation ($\mathbf{b}$), unless otherwise stated, vectors are assumed to be row vectors. We will use bold upper case letters with superscripts indices to refer to different layer Matrix in a network ($\mathbf{W}^1, \mathbf{W}^2$).

This project is organized as follows. In Chapter \ref{chap2} we give an overview of Natural Language Processing and discuss some of its applications. Chapter \ref{chap3} discusses the neural network embeddings, and present word embeddings matrix available in the public domain and their calculation. Chapter \ref{chap4} gives experimental results on a sentiment analysis task and finally Chapter \ref{chap5} concludes our empirical investigation and provides some possible directions for future research.

%% file: chapter2.tex
\chapter{Overview of NLP and its Applications}\label{chap2}

This chapter will introduce Natural Language Processing (NLP), then Neural Network (NN) as a way to solve NLP and will end by discussing some of it applications. 

\section{Natural Language Processing}

\textbf{Natural Language Processing} (NLP) is an ensemble of techniques to process and analyse natural languages---human languages---by  means of computers. Current applications of natural language processing are include sentiment analysis, smart reply, machine translation, bots and expert assistant. 

In this section we will give an overview of natural language processing, and it link to machine learning.

We cannot talk of NLP without mentioning artificial intelligence which is the big field where NLP can be seen as Sub field.

Artificial Intelligence (AI) is a field in computer science that aims to make computers able to make appropriate decision where required without human implicit interaction. 

Natural Language processing is to Artificial Intelligence what languages (spoken and written) are to human. It is the chief master behind technology like \textit{Siri, Google Assistant, $\ldots$}.

Natural Language processing is considered to be a difficult problem in computer science, this is due to the ambiguity and imprecise characteristics of the human languages. 

The early roots of the field date back to the well known \textit{intellectually fertile period} just after the second world war $(1940's - 1950's)$, since then most NLP system were based on a set of complex hand written rules. However in the $1980s$ NLP was revolutionized by the introduction of \textit{machine learning algorithm}, and it was only in $2010$ that deep Neural Network became widespread in natural language processing. IBM Research has been recognized by many as being responsible for several early successes in this field.

Machine learning can be define as a sub-field of AI that aims to make computers learn from the data.

\begin{figure}
	\centering
	\includegraphics[width=0.7\linewidth]{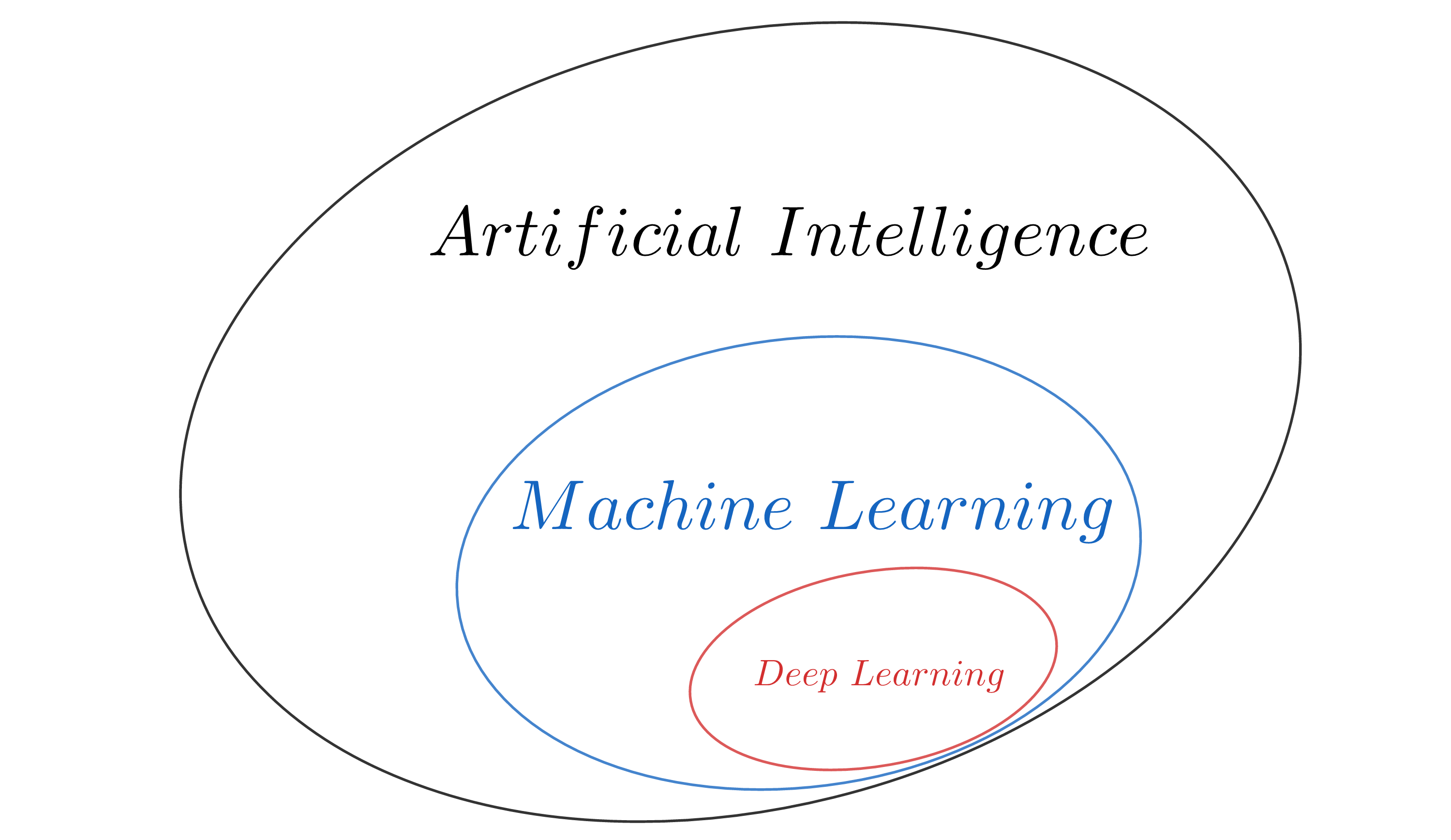}
	\caption{Artificial Intelligence, Machine Learning and Deep Learning}
	\label{fig:aimldl}
\end{figure}

The introduction of deep neural network in NLP caused a big shift in the growth for the field and is the reason of the current trend and amazing application of the output like speech recognition, automatic email reply, bot-expert-agent and much more. See Figure \ref{fig:aimldl} for the relation between \textbf{AI}, \textbf{ML} and \textbf{Deep Learning}. 
The next subsection, will introduce four major machine learning tasks. 

\subsection{Supervised Learning}
Supervised learning is a machine learning task that consists of giving to a model several examples---feature(input) and label(output)---in a way that the model can generalize to unseen data. 

Lets $\left\{\left(x_{1}, y_{1}\right), \ldots,\left(x_{n}, y_{n}\right)\right\}$ be a set of $n$ examples---training data. Supervised learning consist of feeding a model $f$, by feeding we mean roughly giving $f$ examples of input$(x_i)$ and output$(y_i)$ such that :
\begin{align*}
	f(x_i) \approxeq y_i\quad \text{for }i\in \{1,2,\cdots,n\} 
\end{align*} 
Let call $f_{fid}$ the function $f$ after been fed. 

The objective is to make $f_{fid}$ able to predict the value of $y_j$ for $j\in \{a,b,\cdots,m\}$ given unseen data---testing data.

Let $\{x_{a},x_{b},\cdots,x_{m}\}$ be unseen data, we would like to know the unknown label $\{y_{a},y_{b},\cdots,y_{m}\}$.

To be able to measure how well our model performs, as a rule of thumb, the training data is split in two parts
\begin{itemize}
	\item \textbf{Training data} : usually $80\%$ of the general training data 
	\item \textbf{Validation data} : the remaining percentage, that will be viewed as unseen data.
\end{itemize} 

The motivation behind spitting the general training data in two part is to be able to have known $label(output)$ that can be used to evaluate how well $f_{feed}(x_i)$ differ from $y_i$ for $i\in\text{Evaluation data}$.

There exists several machine learning algorithms available for supervised learning approach including neural networks, which one to use is often an empirical question as this depend on the data at hand. 

\subsection{Unsupervised Learning}
Unsupervised Learning is a machine learning task also known as self-organization that consist of making use of unlabeled data. Contrary to the supervised approach, this approach have a set of inputs without output and we build a model to learn internal representation of those inputs in order to make predictions.

Unsupervised task is known to be more difficult---less accurate---than the supervised one but its attractiveness is due to the fact that most available data is unlabeled.   

Some of the most used unsupervised learning include Clustering, Anomaly detection and Neural Networks. 

\subsection{Semi-supervise Learning}
Semi-supervised learning is a class of machine learning task that makes use of unlabeled data for training jointly with small amounts of labeled data. It falls between the two previous machine learning tasks (\textit{supervised learning and unsupervised learning}).

As a rule of thumb it has been found by machine learning researches that \textit{adding} small amount of labeled data to  unsupervised learning improves considerably its accuracy.

Natural Language researchers have been mostly interested on unsupervised and semi-supervised learning as most available textual data are usually unlabeled.

\subsection{Reinforcement Learning}  
Reinforcement Learning can be seen as a special case of supervised learning. This task aims to train an agent to learn to perform actions in an environment so as to maximize some notion of cumulative \textit{reward} over penalties.

\section{Artificial Neural Networks}
Artificial Neural Networks(ANN) or simply Neural Network(NN) are a set of models that are used in machine learning to process information and learn pattern the way a biological brain does.

Artificial refers to the main field Artificial Intelligence, which work on making computers intelligent. contrary to explicitly coding action, the subfield of A.I called machine learning consists of giving data to a model and allow it to learn from it and act as expected. For example a machine learning model can learn how to detect a face in a picture by giving it several examples of facial images. 

The current progress of A.I is due to the contribution of ideas from several disciplines. Artificial Neural Network as a part of Artificial Intelligence comes to life by jointly contribution of two fields Biology and Mathematics.

It has always been the dreams of A.I researchers to build a system that can learn and act independently of human interaction, to achieve this goal scientist started studying the brain, we are still far from the target but the idea of Artificial Neural Network can be considered as one of the results of those investigation. 

\begin{figure}[H]
	\centering
	\includegraphics[width=0.6\linewidth]{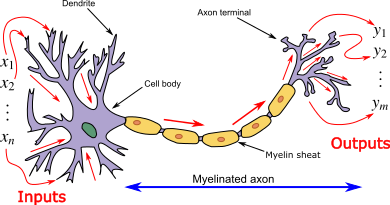}
	\caption{Neuron and myelinated axon, with signal flow from inputs at dendrites to outputs at axon terminals (\textit{Sources: Wikipedia, Neural Network})}
	\label{fig:annwiki}
\end{figure}

Briefly speaking neural networks are the organization of neurons (similar to the neurons in the brain) with connections to other neurons. These neurons pass a message or signal to other neurons based on the received input and form a complex network that learns with some feedback mechanism \citep{Goldberg:2016:PNN:3176748.3176757}. So one may ask, how are the neurons connected? What kind of message do they pass to each other? What do we mean by feedback mechanism? 

Next few lines will give a more detail to answer those questions, but before lets introduce the simplest Neural Network.

\subsection{Perceptron}
The simplest Neural Network is the \textit{perceptron}, which can be written mathematically as :
\begin{align}
	Perceptron(\mathbf{x}) = \mathbf{xW}+\mathbf{b},
	\label{NNP}
\end{align}
where $\mathbf{W}\ :\ \text{Weight matrix},\ \mathbf{b}\ :\ \text{bias term},\ \mathbf{x}\in \mathbb{R}^{d_{in}},\ \mathbf{W}\in\mathbb{R}^{d_{out\times d_{in}}},\ \mathbf{b}\in \mathbb{R}^{d_{out}}$.

\begin{figure}
\centering
\includegraphics[width=.9\linewidth]{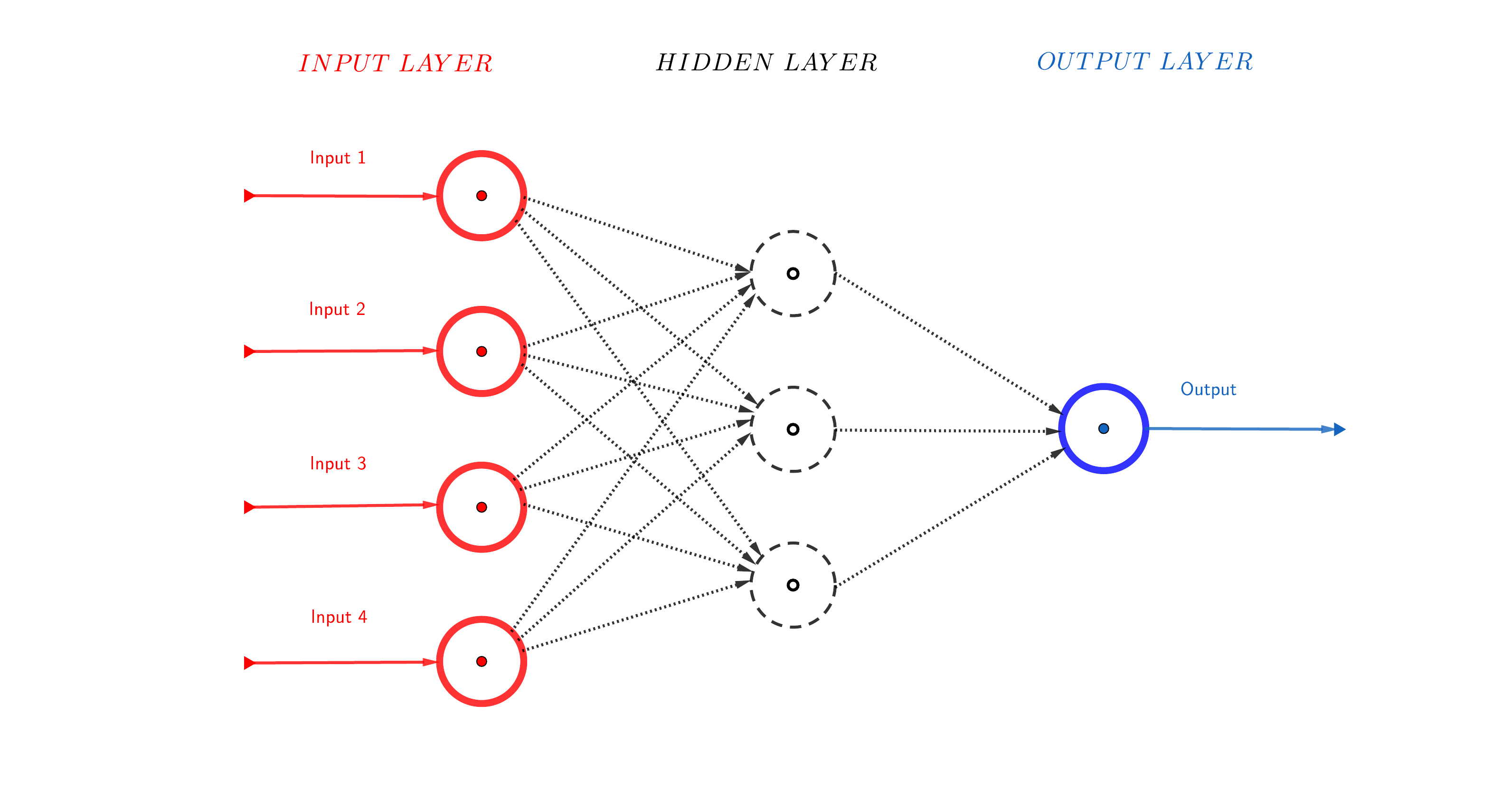}
\caption{Fully Connected Neural Network}
\label{fig:fullyconnectedNN}
\end{figure}%

As you may have noticed Equation \ref{NNP} is exactly like the linear regression equation, consequently this equation can only approximate linear function---a class of those model are known as \textbf{linear models}. 

In real life we usually deal with complex problems that are mostly non linear. We will next introduce a more advance model that can deal with such complex problems. 

\subsection{Feed-forward Neural Network}
First introduced by Frank Rosenblatt in his paper “The perceptron: a probabilistic model for information storage and organization in the brain”\citep{Rosenblatt58theperceptron:} \textit{feed-forward neural network}, also called multilayer perceptron (MLP) is a mathematical function $f$ mapping some set of input values $x$ to output values $y$ and learns the value of the parameters $\theta$ that result in the best function approximation. 
\begin{align}
	y&=f(x, \theta)
\end{align}

The network in Figure \ref{fig:fullyconnectedNN}, is an example of a \textit{feed-forward Neural Network} and the message and the connections between neuron can be written mathematically as follows :

\begin{align*}
MLP(\mathbf{x}) = \mathbf{g}\left(\mathbf{g}(\mathbf{xW^1}+\mathbf{b^1})\mathbf{W^{out}}+\mathbf{b^{out}}\right)
\end{align*}
were :$\mathbf{x}\in \mathbb{R}^{d_{in}},\ \mathbf{W^1}\in\mathbb{R}^{d_{1}\times d_{in}},\ \mathbf{b^1}\in \mathbb{R}^{d_{1}}\ \mathbf{W^{out}}\in\mathbb{R}^{d_{out}\times d_{1}},\ \mathbf{b^{out}}\in \mathbb{R}^{d_{out}}$

The main difference with the previous model is the introduction of a function known as \textbf{activation function} (\textit{also called link/decision/transfer function}).The reason of having such a function $\mathbf{g}$ is to allow the neural network to learn non-linearity. We next introduce and present some common activation functions.

\subsection{Sigmoid}
\begin{figure}
	\centering
	\begin{subfigure}[b]{0.40\textwidth}
		\includegraphics[width=\linewidth]{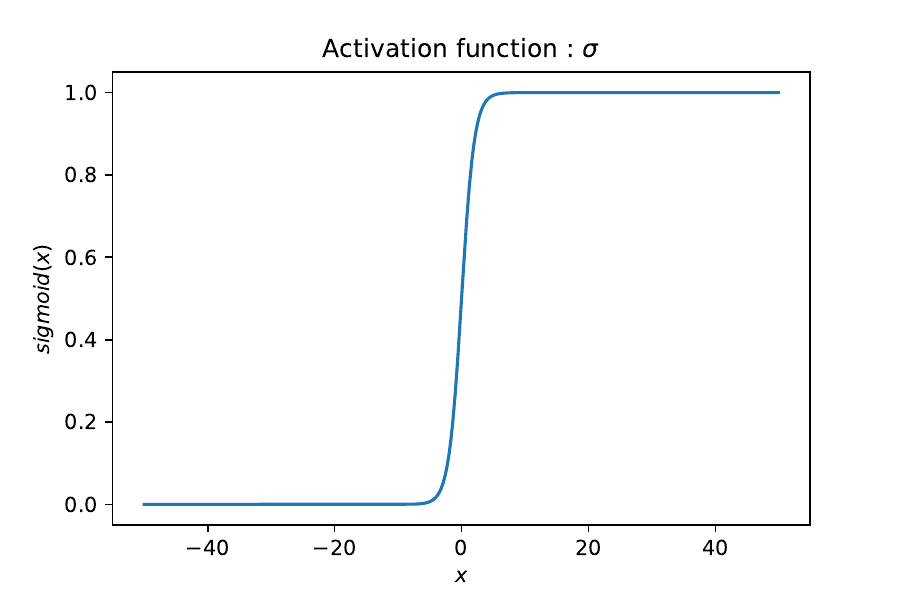}
		\caption{Activation function : \textbf{Sigmoid} ($\sigma$)}
		\label{fig:sigmoid}
	\end{subfigure}%
	\begin{subfigure}[b]{0.4\textwidth}
		\includegraphics[width=\linewidth]{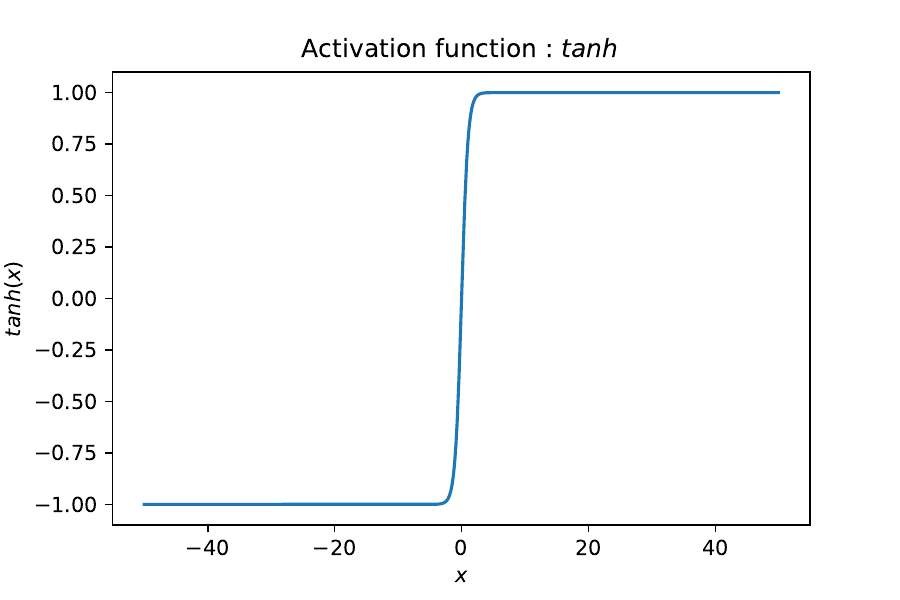}
		\caption{Activation function : \textbf{tanh}}
		\label{fig:tanh}
	\end{subfigure}\\
	\begin{subfigure}[b]{0.4\textwidth}
		\includegraphics[width=\linewidth]{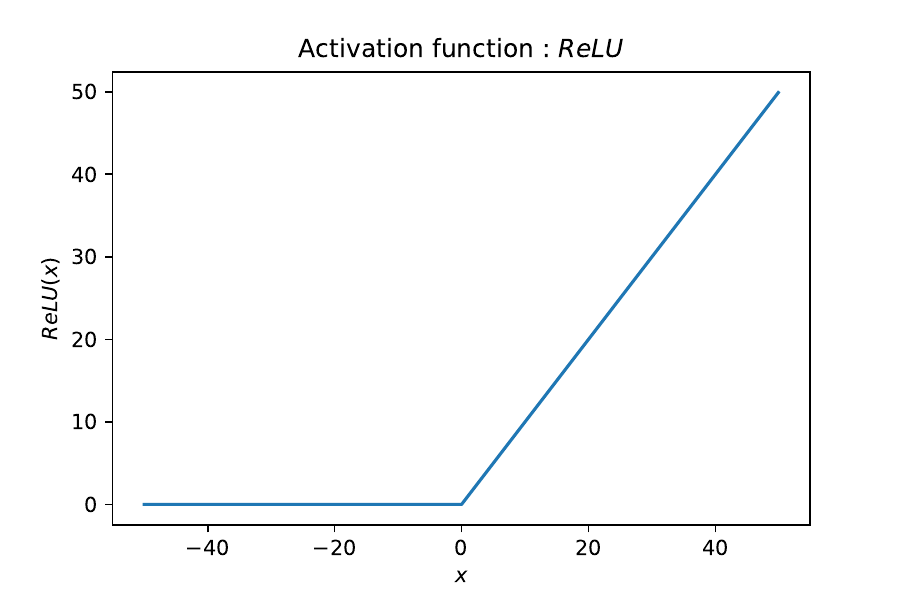}
		\caption{Activation function : \textbf{ReLU}}
		\label{fig:Relu}
	\end{subfigure}%
	\begin{subfigure}[b]{0.4\textwidth}
		\includegraphics[width=\linewidth]{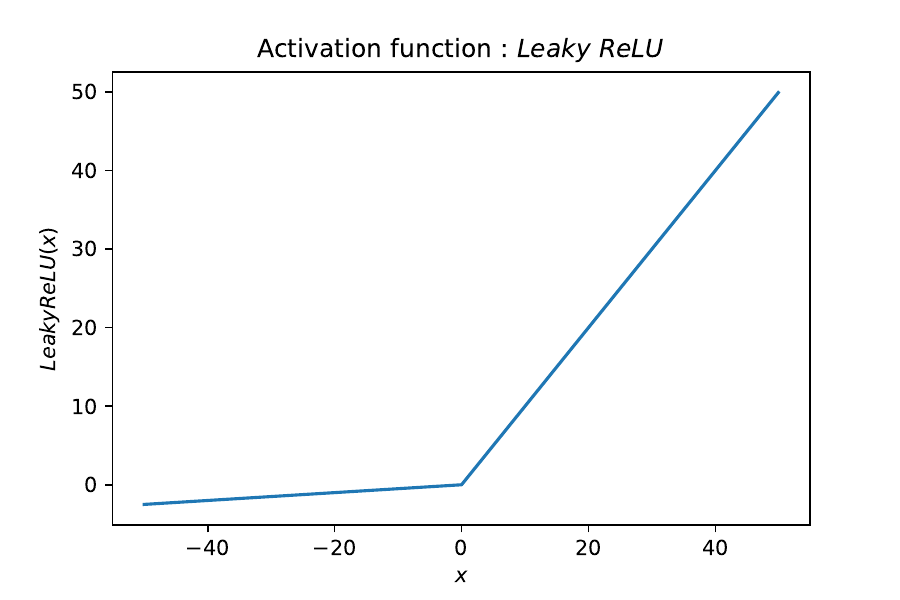}
		\caption{Activation function : \textbf{Leaky ReLU}}
		\label{fig:Leaky_ReLU}
	\end{subfigure}
	\caption{Common Activation functions, the choice of which one depend on the task at hand (range of outputs)}
\end{figure}

(see Figure \ref{fig:sigmoid})
The sigmoid activation function (also called logistic function) is define as
\begin{equation*}
\sigma(x)=1/(1+e^{-x}),
\end{equation*} 

The sigmoid activation function is an s-shaped function, transforming each $\mathbf{x}$(input) value into the range $[0, 1]$.

This activation function has a good propriety of being smooth and differentiable $\sigma^{\prime}(x)=\sigma(x)(1-\sigma(x))$, which are very useful for the feedback mechanism that we will explain later in this work

\subsection{Hyperbolic tangent (tanh)}(see Figure \ref{fig:tanh})

The hyperbolic tangent activation function defined as
\begin{equation*}
	tanh(x)=\frac{e^{2x}-1}{e^{2x}+1},
\end{equation*}
is an s-shaped function, transforming the values x into the range $[-1, 1]$. NLP tasks mostly rely on the tanh activation function.

\subsection{ReLU}(see Figure \ref{fig:Relu})
The Rectifier activation function \cite{Glorot2011DeepSR} is a more recently popularized activation function, also known as the rectified linear unit or ramp function, is a very simple activation function that is easy to work with and was shown many times to produce excellent results. It has been proved the default recommendation for modern Neural Network calculations \cite{glorot10a_relu}. 

In addition ReLu is a special activation function, as it can be seen as two joined linear functions, this makes it inherit some of the useful propriety of linear models that make it generalize well.

The ReLU unit clips each value $x < 0$ at $0$. 

\begin{align*}
ReLU(x)=max(0, x)&=
\left\{
\begin{array}{cl}
0&x<0\\
x&\text{otherwise}
\end{array}
\right.
\end{align*}

As a rule of thumb, \textbf{ReLU} units work better than \textbf{tanh}, and \textbf{tanh} works better than \textbf{sigmoid}.

\subsection{Leaky ReLU}(see Figure \ref{fig:Leaky_ReLU})
The horizontal line with $0$ in ReLU as the output which is a constant with a derivative of $0$ and therefore may become a bottleneck during training, as the weights will not easily get updated. To solve the problem, there was a new activation function proposed: Leaky ReLU, where the negative value outputs a slightly slanting line instead of a horizontal line, which helps in updating the weights through back-propagation effectively. 

Leaky ReLU  is defined as

\begin{align*}
LeakyReLU&=
\left\{
\begin{array}{cl}
x&\text{if : }x>0\\
\alpha x& \text{if : } x<0
\end{array}
\right.
\end{align*}

- with $\alpha$ a parameter that is defined as a small constant, say 0.005

The network in Figure \ref{fig:fullyconnectedNN} has only one hidden layer with only three nodes, those nodes are called artificial neurons in comparison to neurons in a biological brain. This kind of simple neural network are limited as to the generalization power. 
\begin{figure}
	\centering
	\includegraphics[width=0.5\linewidth]{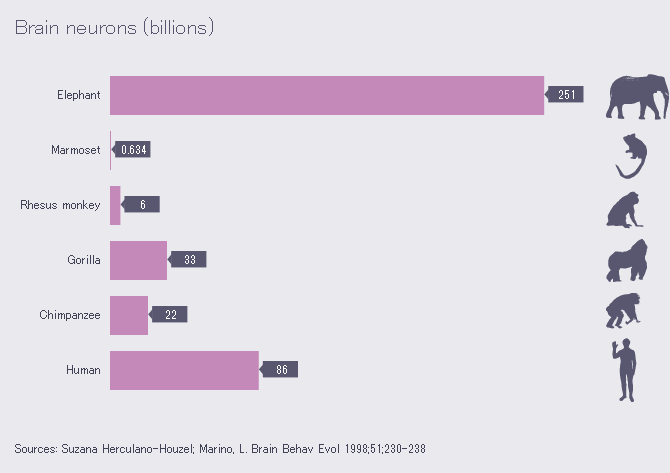}
	\caption{Comparison of number of neurons (billions). Elephant, Marmoset, Rhesus monkey, Gorilla, Chimpanzee and Human. \textit{Sources: Suzana Herculano-Houzel}}
	\label{fig:brainsizecomparison-brainneuronsbillions}
\end{figure}
In Figure \ref{fig:brainsizecomparison-brainneuronsbillions} we can see how huge the size of neuron connection in biological creatures can be, in parallel modern Neural Network require much bigger (number of layers) and more deeper (number of neurons) Neural Network. Now one may ask, how big ? how deep ? this is basically an empirical question as there is no theory providing an answer to these question but at the theoretical point, the Universal Approximation Theorem (\textbf{UAT}) states that a Feed-forward Networks with a single hidden layer containing a finite number of neurons can approximate continuous functions on compact subsets of $\mathbb{R}^n$, under mild assumptions on the activation function.

Flat structures with only a small number of layers, may not be able to compute complex problems without using a very large and difficult to handle number of hidden nodes which lead to a whole field call Deep Learning \citet{Goodfellow-et-al-2016}.

As discussed in the previous section shallow neural networks have limited generalization power. In the next section we will talk about Deep Neural Networks and discuss finally the mechanism of training a neural network cited at the introduction of this chapter. 
\section{Deep Neural Networks}
Even tough it was only in $1986$ that the term Deep learning was introduced to the machine learning community by Rina Dechter, it only came to artificial neural network  in $2000$ by Igor Aizenberg.

\begin{figure}
	\centering
	\includegraphics[width=0.7\linewidth]{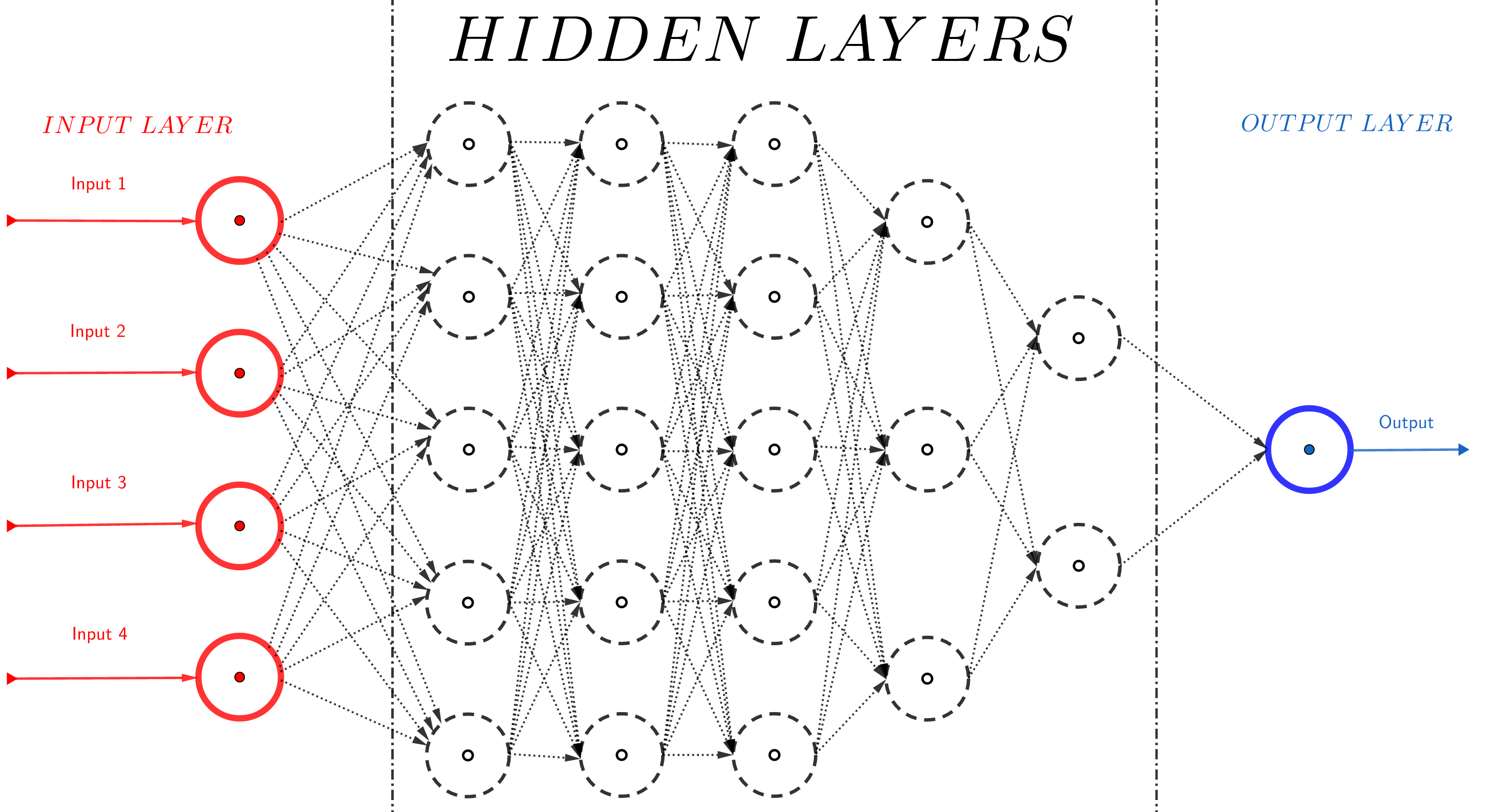}
	\caption{Deep Neural Network with \textit{four} hidden layers, note that we have not included the bias neurons}
	\label{fig:deepneural}
\end{figure}

A Deep Neural Network is simply an artificial neural network with several hidden layers. See Figure \ref{fig:deepneural}. 

In this section we will give details on training a deep neural network. Before reaching there, we will give a specific example. Figure \ref{fig:deepneural} represents an example of a deep neural network with $5$ hidden layers. 

Note that $\mathbf{g}^i$ denotes the activation function used in layer $i$, this includes the identity activation function.   

The input data has the form $\mathbf{x}\in \mathbb{R}^4$, parameters---weights and biases---in layer 2---layer 1 in the input layer--- is represented by a weight matrix $\mathbf{W}^2 \in \mathbb{R}^{5\times 4}$ and a vector $b^2\in \mathbb{R}^5$, respectively. The output of this layer has the form 
\begin{equation}
	\mathbf{g}^2\left(\mathbf{W}^2 x + \mathbf{b}^2 \right)\in \mathbb{R}^5
	\label{layer_2}
\end{equation}

Layer three has $5$ neurons as well, each receiving input from the output of \ref{layer_2}. The weight matrix in layer 3 is $\mathbf{W}^3 \in \mathbb{R}^{5\times 5}$ and the bias is $\mathbf{b}^3\in \mathbb{R}^5$, respectively. We obtain the following output that is transferred to layer four  
\begin{equation}
	\mathbf{g}^3\left(\mathbf{W}^3\mathbf{g^2}\left(\mathbf{W}^2 x + \mathbf{b}^2\right)+\mathbf{b}^3 \right)\in \mathbb{R}^5
	\label{layer_3}
\end{equation}

Layer four has $5$ neurons as well, each receiving input from the output of \ref{layer_3}. The weight matrix in layer four is $\mathbf{W}^4 \in \mathbb{R}^{5\times 5}$ and the bias is $\mathbf{b}^4\in \mathbb{R}^5$, respectively. We obtain the following output that is transferred to layer five  
\begin{equation}
\mathbf{g}^4\left(\mathbf{W}^4\mathbf{g}^3\left(\mathbf{W}^3\mathbf{g^2}\left(\mathbf{W}^2 x + \mathbf{b}^2\right)+\mathbf{b}^3 \right)+ \mathbf{b}^4\right)\in \mathbb{R}^5
\label{layer_4}
\end{equation}

Layer five has $3$ neurons, each receiving input from the output of \ref{layer_4}. The weight matrix in layer five is $\mathbf{W}^5 \in \mathbb{R}^{3\times 5}$ and the bias is $\mathbf{b}^5\in \mathbb{R}^3$, respectively. We obtain the following output that is transferred to layer six  
\begin{equation}
\mathbf{g}^5\left(\mathbf{W}^5\mathbf{g}^4\left(\mathbf{W}^4\mathbf{g}^3\left(\mathbf{W}^3\mathbf{g^2}\left(\mathbf{W}^2 x + \mathbf{b}^2\right)+\mathbf{b}^3 \right)+ \mathbf{b}^4\right)+ \mathbf{b}^5\right)\in \mathbb{R}^3
\label{layer_5}
\end{equation}

The sixth Layer has $2$ neurons, each receiving input from the output of \ref{layer_5}. The weight matrix in layer six is $\mathbf{W}^6 \in \mathbb{R}^{2\times 3}$ and the bias is $\mathbf{b}^6\in \mathbb{R}^2$, respectively. We obtain the following output that is transferred to layer seven (\textit{output}) 
\begin{equation}
\mathbf{g}^6\left(\mathbf{W}^6\mathbf{g}^5\left(\mathbf{W}^5\mathbf{g}^4\left(\mathbf{W}^4\mathbf{g}^3\left(\mathbf{W}^3\mathbf{g^2}\left(\mathbf{W}^2 x + \mathbf{b}^2\right)+\mathbf{b}^3 \right)+ \mathbf{b}^4\right)+ \mathbf{b}^5\right)+ \mathbf{b}^6\right)\in \mathbb{R}^2
\label{layer_6}
\end{equation}

The Seventh Layer has $1$ neurons, which is our last layer---output layer, note that the number of neurons on the output layer determine which kind of problem we are dealing with \textbf{regression} or \textbf{classification}. The weight matrix in layer seven is $\mathbf{W}^7 \in \mathbb{R}^{1\times 2}$ and the bias is $\mathbf{b}^7\in \mathbb{R}^1$, respectively. We obtain the following output, which is the total output of our model $M(x)$ (\textit{output}) 
\begin{equation}
M (x)=\mathbf{g}^7\left(\mathbf{W}^7\mathbf{g}^6\left(\mathbf{W}^6\mathbf{g}^5\left(\mathbf{W}^5\mathbf{g}^4\left(\mathbf{W}^4\mathbf{g}^3\left(\mathbf{W}^3\mathbf{g^2}\left(\mathbf{W}^2 x + \mathbf{b}^2\right)+\mathbf{b}^3 \right)+ \mathbf{b}^4\right)+ \mathbf{b}^5\right)+ \mathbf{b}^6\right)+ \mathbf{b}^7\right)\in \mathbb{R}
\label{layer_7}
\end{equation}

View globally the network in Figure \ref{fig:deepneural} is equivalent to the Equation \ref{layer_7}, which is defined as $M:\mathbb{R}^4\rightarrow\mathbb{R}$ in terms of its $114$ parameters---total number of weight and bias vectors.

To generalize the concept let suppose the network has $L$ layers, with layer $1$ being the input layer and the layer $L$ being the last layer---output layer. Let $n_i$ be the number of neurons in layer $i$. So $n_1$ and $n_L$ is the dimension of the input data and the output dimension, respectively. 

The matrix of weights at layer $l$ will be $\mathbf{W}^l\in\mathbb{R}^{n_l\times n_{l-1}}$ , the weight of neuron $i$ at layer $l$ that multiply the output from layer $j$ from layer $l-1$ will be written as $\mathbf{w}^l_{ij}$ and the bias vectors as $b^l \in \mathbb{R}^{n_l}$.

The complete fending of our general model, from the input of dimension $\mathbb{R}^{n_1}$ to the output $\hat{y}^L \in \mathbb{R}^{n_L}$ can be written as 
\begin{align}
	\hat{y}^{1} &= x\in \mathbb{R}^{n_1}\\
	\hat{y}^{l} &= \mathbf{g}\left(\mathbf{W}^{l} \hat{y}^{l-1}+ b^{l} \right)\in \mathbb{R}^{n_{l}},\qquad for\ l=2,3,4,\cdots, L.
\end{align}
With $\mathbf{g}$ been the any activation function, include the identity activation function.

Suppose $x$ of shape $(N, n_1)$ be our input and $y$ of shape $(N, n_L)$ our target outputs, with $N$ the number of examples in our data. Training our model over such input $x$, can be viewed as an optimization problem. The quadratic cost function that we would like to minimize has the form

\begin{equation}
	Cost = \frac{1}{N}\sum_{i=1}^{N}\frac{1}{2}||y_i-\hat{y}^L(x_i)||^{2}_{2},\label{cost_funct}
\end{equation}

\subsection{Training a Neural Network}
Training a NN, consist of adjusting parameters, in order to reduce the Cost function \ref{cost_funct}. The aims when training a neural network, is to increase his representation power---the ability to predict unseen data.

\begin{figure}
	\centering
	\includegraphics[width=0.7\linewidth]{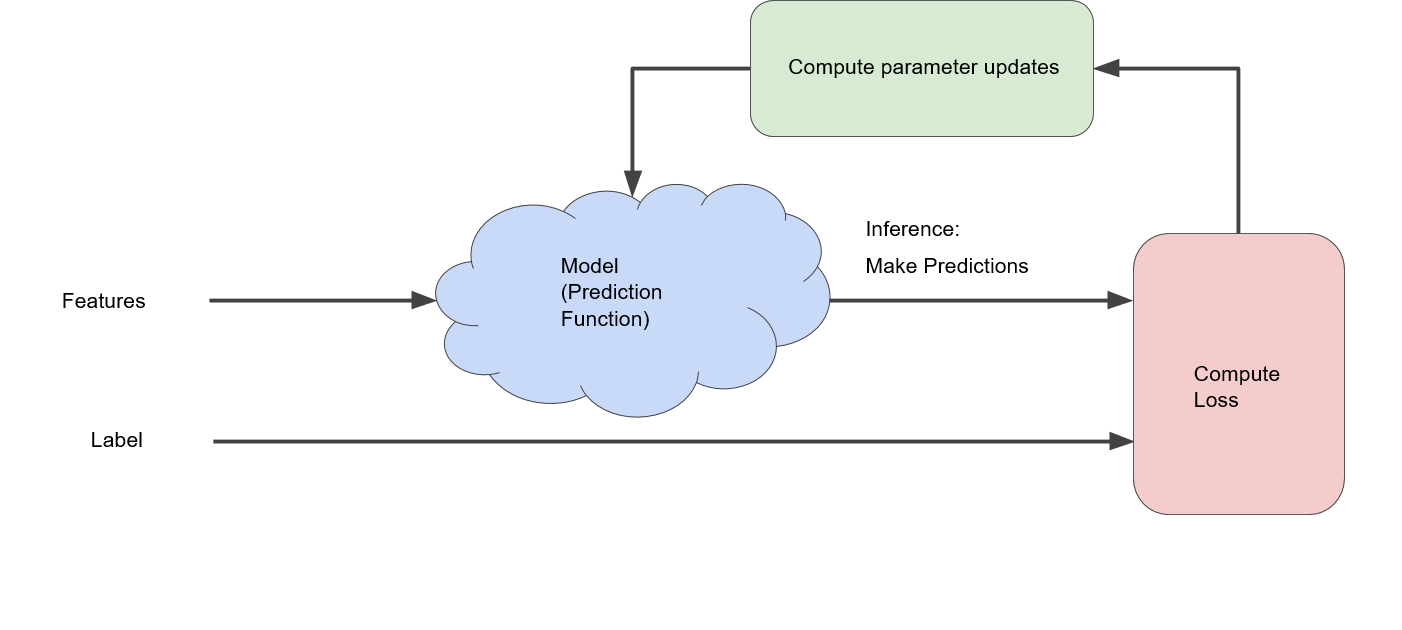}
	\caption{Gradient Descent Mechanics \citep{crashcourse}}
	\label{fig:gradientdescentdiagram}
\end{figure}

Roughly speaking, all training methods work by repeatedly computing an estimate of the error over the dataset, computing the gradient with respect to the error, and then moving the parameters in the direction of the gradient \cite{Goldberg:2016:PNN:3176748.3176757}.

Sometimes training a neural network can be time consuming, specially if it is a deep neural network.

The so called \textit{big bang of deep learning}---the introduction of GPU--- has revolutionized the training of a deep neural network as algorithms that needed weeks to be trained could now be done only in few hours. 

The notion of error is at the center of the training of the neural network. You mostly encounter the term \textbf{Evaluation function} and \textbf{Loss function} when dealing with training neural networks.

\subsection{Loss function vs Evaluation function}
\textbf{The loss function} is applied to the output of the network during training to computer error between the predict value for a particular \textit{example} and the true value(target).

This error function is used in training. An example of it is the \textit{squared error} for a regression problem or cross-entropy for classification. 

The square error is defined as :
\begin{align}
	SE = (y, f(x, \theta))= (y - \hat{y})^2,
\end{align}
where $y :$ the true output also called  and $\hat{y}=f(x, \theta)$ the predicted value of $y$ by our NN $f$ with parameters $\theta$

\textbf{The evaluation function} on the other hand is applied to the entire network, it is the really error we want to improve, an example of could be \textit{accuracy} for a classification problem.

The cross-entropy loss used in classification problem is defined by :
\begin{align}
CE = (y, f(x, \theta)=\hat{y})= \left((1-y)\log(1-\hat{y}) + y\log(\hat{y})\right)
\end{align}
with $y :$ the true output also called target and $\hat{y}=f(x, \theta)$ the predicted value of $y$ by our NN $f$ with parameters $\theta$.

There are many algorithms for training a NN, models differs in the way the error is computed, and how back-propagation is used. In this section, we describe the stochastic gradient descent (SGD), which is the basic algorithm as others can be seen as a modified version of it. 


\subsection{Gradient Descent and Back-Propagation}
Let $f(x, \theta)$ be our neural network. Training $f(x, \theta)$ consist of tuning the value of $\theta$ such that $f(x)=y\approxeq f(x, \theta)(x, \theta)=\hat{y}$.

The gradient descent is method that consist of computing the derivative of the loss function relative to the parameters $\theta$, by parameters we mean \textit{weights} and \textit{biases} of the neural network.


Back-propagation basically finds the gradients, and the gradient descent uses those gradients to update the network parameters $\theta$ \citep{book:2313503}. See Figure \ref{fig:gradientdescentdiagram} for a visual description. 

Gradient descent is an optimization algorithm for finding the minimum of a function. 

The back propagation algorithms is a family of methods that use gradient descent---find the minimum of a function approach--- by calculating the gradient of the loss function. 

The algorithm below describe how the stochastic gradient descent works :
\begin{algorithm}[H]
	\caption{Online Stochastic Gradient Descent Training}\label{euclid}
	\begin{algorithmic}[1]
		\State \textbf{Input}:  Function $f(\mathbf{x}; \theta)$ parameterized with parameters $\theta$.
		\State \textbf{Input}:  Training set of inputs $\mathbf{x}_{1}, \ldots, \mathbf{x}_{\mathbf{n}}$ and outputs $\mathbf{y}_{1}, \ldots, \mathbf{y}_{\mathbf{n}}$
		\State \textbf{Input}: Loss function $L$
		\While{stop criteria not met} \State{Sample a training example $\mathbf{x_i}, \mathbf{y_i}$.}
		\State{Compute the loss $L(f(\mathbf{x_i}; \theta), \mathbf{y_i})$}
		\State{$\hat{\mathbf{g}} \gets$ gradients of $L(f(\mathbf{x_i} \theta), \mathbf{y_i})$ w.r.t $\theta$}
		\State{$\theta\gets \theta + \eta_{k} \hat{\mathbf{g}}$} \EndWhile
		\State \Return $\theta$
	\end{algorithmic}
\end{algorithm}

The back propagation algorithm basically as shown in the above algorithm, decides how the value of $\theta$ changes after comparing $f(\mathbf{x_i},\theta)$ (the predicted outputs) with $y_i$ (the real outputs). To accomplish this it computes the derivative of the error with respect to the parameters of the model ($\theta$) $\displaystyle \frac{d E}{d w_{i j}}$ and $\theta$ is updated accordingly. 

\begin{figure}
	\centering
	\includegraphics[width=.7\linewidth]{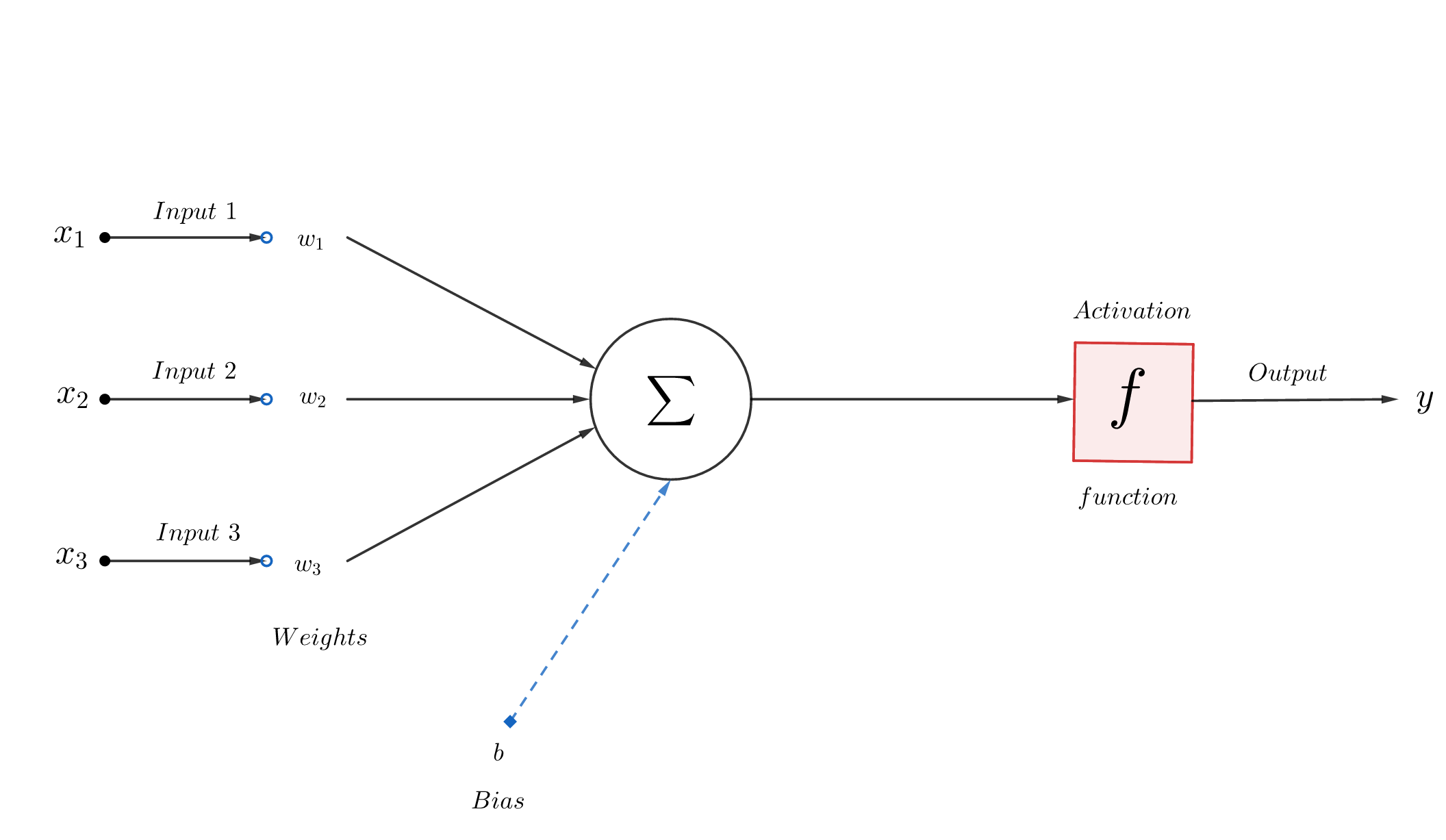}
	\caption{Artificial neuron}
	\label{fig:ArtNeu}
\end{figure}

Let illustrate the above training algorithm by applying it to the single unit on Figure \ref{fig:ArtNeu}.

The output to this unit can be written as :
\begin{equation*}
	y=\sigma(u),
\end{equation*}
where $u = \mathbf{W}^T\mathbf{x}+b$ and $\sigma(x)=\frac{1}{1+e^{-x}}$. 

We have decided to take the sigmoid function as the activation function to make derivation and explanation straightforward for our illustration.

We will use the mean square error(MSE) as our error function :
\begin{equation*}
	E=\frac{1}{2}(\hat{y}-y)^{2}
\end{equation*}

As required by the stochastic gradient descent, the derivative of $E$ with regard to $\mathbf{w}_i$,
$$
\begin{aligned} 
	\frac{\partial E}{\partial \mathbf{w}_{i}}=\hat{\mathbf{g}} &=\frac{\partial E}{\partial y}t \frac{\partial y}{\partial u}  \frac{\partial u}{\partial w_{i}} \\ 
	&=(y-\hat{y})  y(1-y) x_{i} 
\end{aligned} 
$$

Finaly, $\theta$ which is equivalent to $\mathbf{w}$ will be updated in the following way :

\begin{equation}
	\theta=\mathbf{w}^{new}=\mathbf{w}^{old}-\eta\cdot\hat{\mathbf{g}}
\end{equation}

The same logic follows for bigger networks. 

One of the problems that usually occur when the model start adjusting itself, when there is noises in the data the model start adjusting itself on the noises.

So what is noise ? Let understand it by this example : one may have been given data that shows when Salomon decides to go for supper and from the data it is clear that he always goes around 6:30. with this information the model can make a rule that say \textbf{Salomon goes for supper every 6:30}, but lets say on a Saturday when Salomon went to play soccer, one of his friends got injured and as they went to the hospital, Salomon comes for supper at 9:00, the model will update its rule like \textbf{Salomon go for supper every 6:30 except on Saturday} this is wrong as this event was unexpected of Salomon's behaviour and can be considered as \textbf{noise}.

So now, how can our model, know whether a pattern is real or a noise ? This question lead to the notion of regularization. 
 
\subsection{Regularization}

\begin{defn}[Regularization]
	Basically Regularization is a term that describes attempts to avoid overfitting, which happen when the our model accurately predict on the trained data that on the unseen, in other word the model started to memorizing instead of generalizing.   
\end{defn}

There are several approach to this, but we will only present some of them.

\begin{itemize}
	\item Early stop
Early stop is usually refer in literature like regularization in time, the idea in this technique is to stop training when our model start over complicating itself. 

The idea of simple model goes as far as the first machine learning theoretician William of Ockham from 13th century, who developed the idea of Ockham's Razor, basically saying that a model should be as simple as possible.

\textit{The less complex a Machine Learning model, the more likely that a good empirical result is not just due to the peculiarities of the sample.}

The fundamental tension of machine learning is between fitting our data well, but also fitting the data as simply as possible.

Early stop is motivated by this idea and practically, this technique tracks how well our model generalizes on unseen data usually a validation set and when there is no improvement anymore we  stop the training process early. 

%

\item \textbf{Dropout Regularization} : a technique that consist of dropping some nodes(depending on the parameter) in the model at random, to reduce computation and usually help in decreasing over-fitting. See Figure \ref{fig:deepneuraldropout} for visual illustration.  
\begin{figure}
	\centering
	\includegraphics[width=0.7\linewidth]{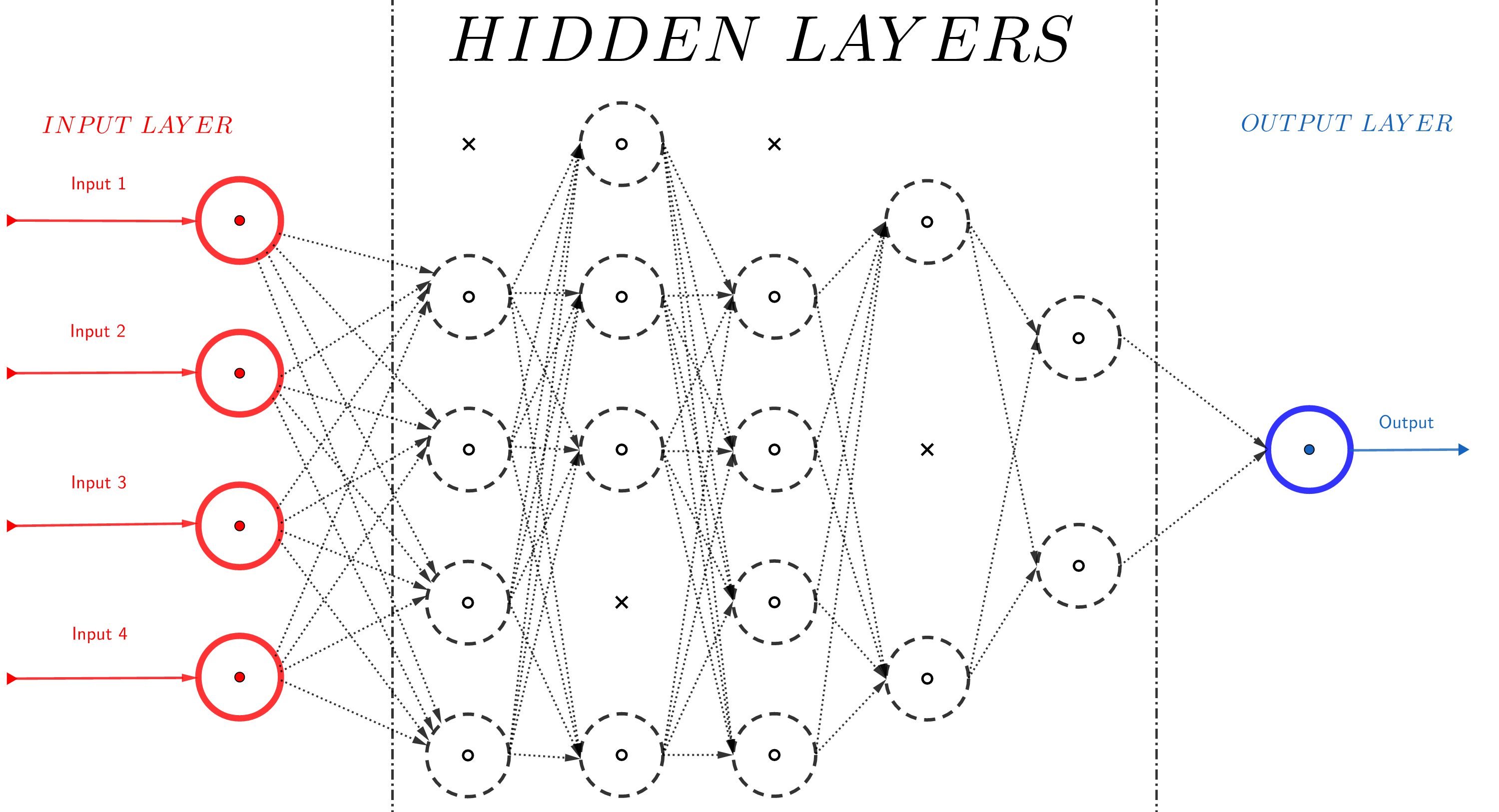}
	\caption{Deep Neural Network in Figure \ref{fig:deepneural} after applying dropout technique}
	\label{fig:deepneuraldropout}
\end{figure}

\end{itemize}

To end this chapter, we will survey some popular application (task) of natural language processing and induce as well the main application task of this essay.
\section{NLP Application}
Applications to NLP are much more closer to our daily life, and familiar to us than we can imagine, from the moment you search for the best restaurant in town on google using Siri, using spell checking in Microsoft Word to correct your spelling, translating a sentence from English to french you are already using the power of NLP.

Next sub-sections will review some common task in NLP.
\subsection{Part-Of-Speech Tagging (POS)}
This task aims to give unique tag to each word as to know its syntactic role. 

A same word can have different syntactic role, for example, in the sentence \textit{the book in the library}, the word \textit{book} will be tagged as noun, but the same word can be tagged as verb in the sentence \textit{to book an interview}. 

\subsection{Question answering}
This task consist of building a model that when given a question can provide an answer. current research work consist of trying to answer open-ended question such as "what is the meaning of life ?" 

\subsection{Named Entity Recognition (NER)}
This task aims to determine which token in a given text map to proper names. 
Although capitalization can be used in recognizing named Entity, but this convention can be misleading as the first word of sentence is also capitalized and some other languages like Arabic don't have any capitalization. 

\subsection{Speech recognition}
This task consist of obtaining text representation from a human speech.
This task is considered as one of the most difficult task in natural language processing as people speaking the same language can speak a word that sounds completely different depending on their accent.

\subsection{Machine Translation}
This task aims to translate automatically a text from a language $\mathbf{a}$ to a language $\mathbf{b}$. This task is among the category of NLP task called \textit{AI-Complete} as it require to give a machine the general knowledge that human have which enable them to effectively translate a text from a language to an other.

\subsection{Semantically Related Words}
This task aims to predict if two words are semantically related which use the WordNet\footnote{http://wordnet.princeton.edu} data base as ground truth.

\subsection{Sentimental Analysis}
Sentiment analysis also known as opinion mining or multimodal sentiment analysis is a concept that uses natural language processing, in combination with other techniques to systematically identify, extract, quantify, and study effective states of text with goal to predict, extract the subjective attitude---sentiment---on it. This kind of sentiment can be in form of a binary rating---like/dislike---or images emoji expressing someone's sentiment. 

Sentiment analysis can be used by a company for example to analyze tweets to determine if people like or dislike their recently released product, it's often used to conduct online reviews. Products of sentiment analysis have become hugely commercially important. there are a lots of company that scan the internet every hour to make buying and selling decisions.

Today social media and online market has grown in a remarkable way and sentiment analysis can be used for example to identify trends of public opinion about a product and use it to increase productivity.

In the context of this work, we will use the well known Internet movie database(IMdb) that contains $50.000$ reviews with custumers sentiment regarding those movies \citep{dataset:2011:ACL-HLT2011}.

%% file: chapter3.tex
\chapter{Neural Network Embeddings}\label{chap3}
Since the popularity of neural networks, people have been working on taking advantage of this powerful tool to solve specific problems in their field. Word embeddings are the core innovation that has brought neural networks to the forefront of Natural Language Processing. 
\begin{defn}
	Neural word embeddings are distributed representations that use continuous values to represent words in a learned vector space of meaning \cite{phd_gouws2016training}.
\end{defn}

In this chapter we will explain what embeddings are, why are they useful and review various mechanisms that have been proposed for their calculations, but before let discuss feature representation to motivate the importance of word embeddings.

\section{Feature Representation}
Text is the most widespread form of data in Natural Language Processing---we write and read them everyday. In this section we will have a look into approaches on how to transform this kind of data into a form that can be processed by a Neural Networks.
%

Like most Machine Learning model, Neural Networks doesn't take raw text as input, so come the notion of \textit{Vectorizing} witch consist of transforming texts into numerical vectors, there exists many ways of doing this, the subsection will survey the most popular ones.


\subsection{Sparse Word representations (One Hot)}
Generally, people use this techniques, consisting of having a vector which assign one on our target element and zero elsewhere. In this concept each feature have its own dimension, the more unique words you have, the bigger the dimensionality of your vector.

\begin{figure}[H]
	\begin{subfigure}{.5\textwidth}
		\centering
		\includegraphics[width=.95\linewidth]{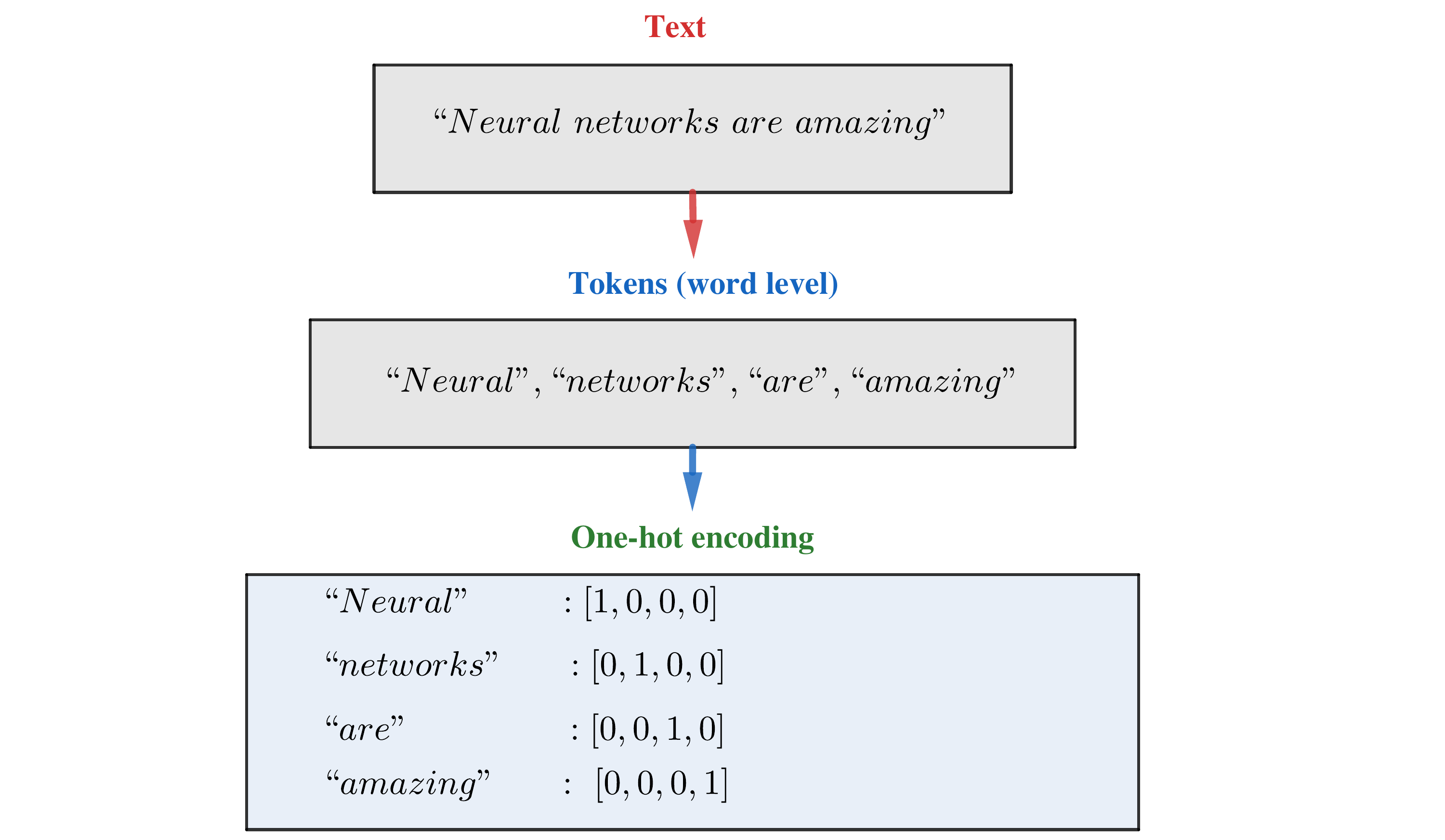}
		\caption{Tokenization, one-hot encoding of the sentence : \textit{"Neural Network are amazing"}}
		\label{fig:one-hot}
	\end{subfigure}%
	\begin{subfigure}{.5\textwidth}
		\centering
		\includegraphics[width=.95\linewidth]{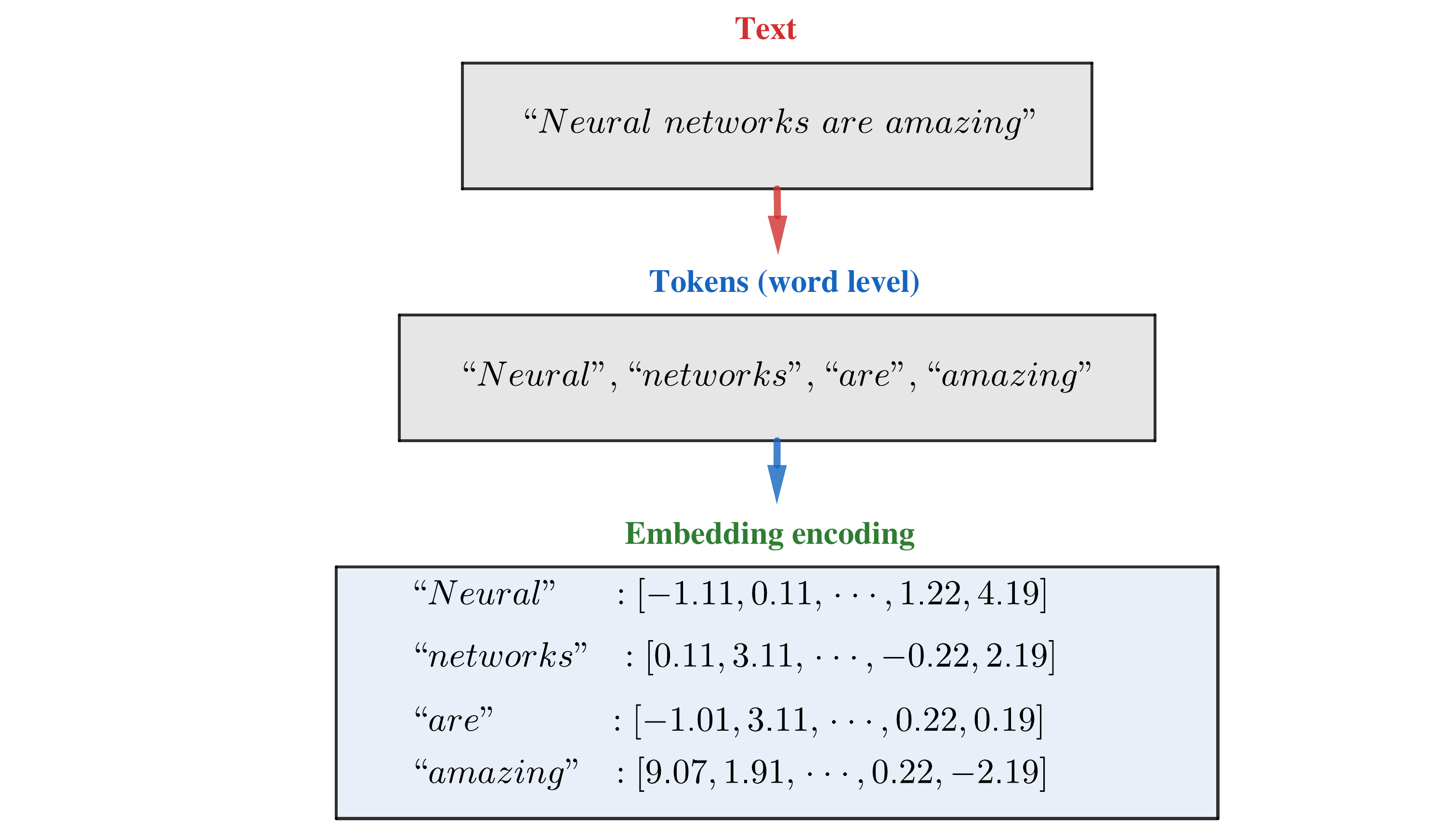}
		\caption{Embedding}
		\label{fig:embedding}
	\end{subfigure}
	\label{fig:one_embedding}
\end{figure}

An example, on how to construct a "one-hot" vector can be seen in Figure \ref{fig:one-hot}.

As you can see form Figure \ref{fig:one-hot} the length of the vector is equal to the size of the total unique words in the corpora. This lead quickly to a big problem as in Natural Language Processing input features can be paragraphs, books, and will definitely have very large number of unique element and this lead to big, sparse vector full of zeros and with only one on the target word that bring no information about the similitude or dissimilitude of features. 

Additionally, words, that are rarely seen during training will be represented poorly. Words that are not seen once during training cannot be classified later and this lead to poor generalization power.

\subsection{Dense Word representations (Token Embedding)}
This concept consists of representing data into a fixed dimensional vector space. 

One of the advantage of using Dense representation is computational, but the main benefit of having a dense representation is the generalization power as feature that are semantically the same will share statistical strength. For example a word like man and women will be more similar than man and study. this property will help our model to infer women when trained in data that contains only the word man.

Token Embedding, also known as \textbf{word embeddings} can be used for a dense representation. Word embeddings are a mapping from discrete objects, such as words, to a dense vectors of real numbers.

To illustrate this approach we are going to used the same example that was use for the \textit{one-hot} vector representation and see how that differs from the token embedding representation. 

\begin{figure}
	\centering
	\includegraphics[width=1\linewidth, height=1\textheight]{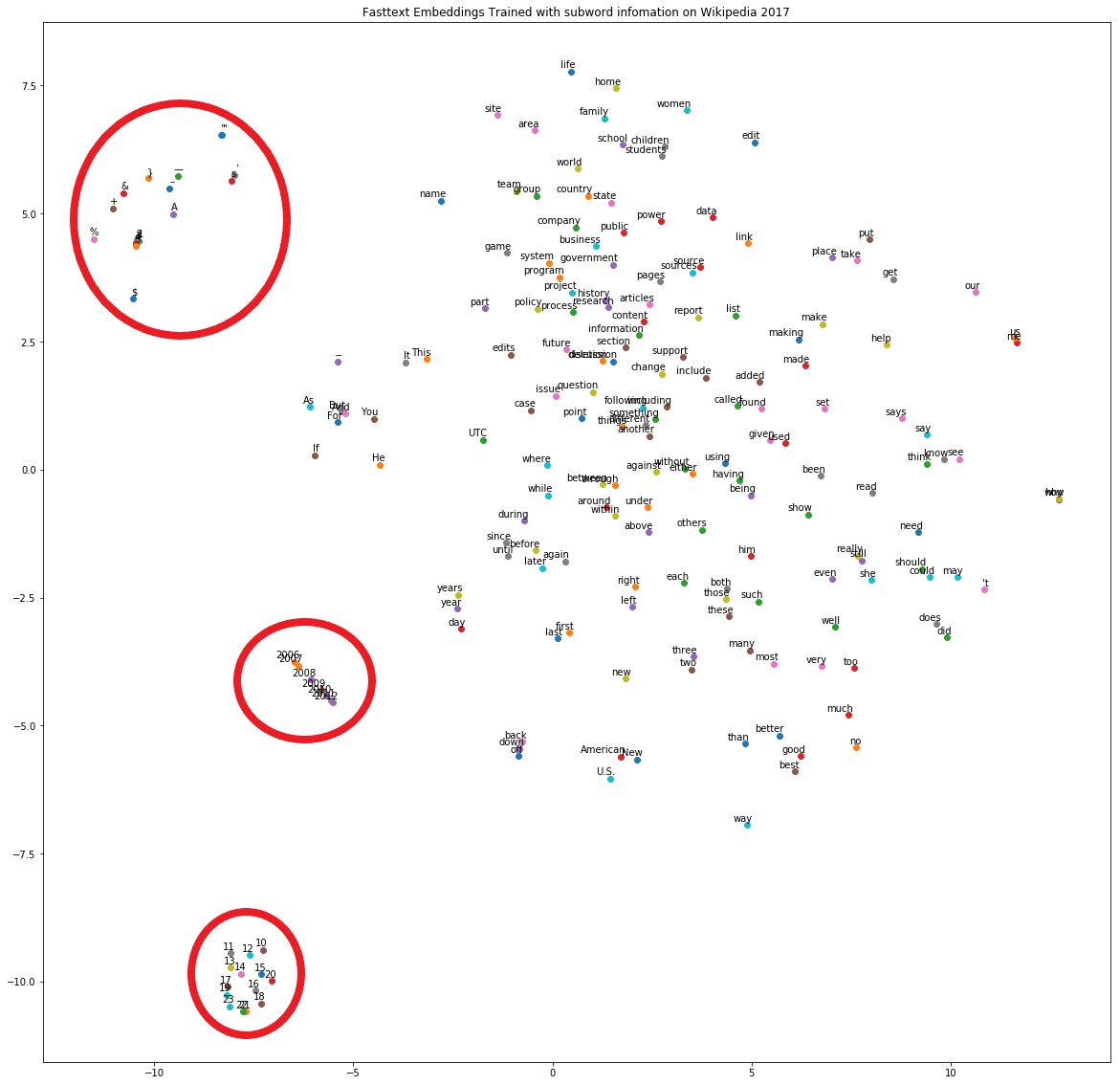}
	\caption{Example of $300$-dimension word embeddings cast in $2$-dimension, for visualization purpose. (similar words are plotted near each other)}
	\label{fig:neuralembpic}
\end{figure}

As you can see from the Figure \ref{fig:embedding}, instead of a sparse vector, we have a dense vector with a much lower representation dimensionality. It common to have embedding dimension between $50$ and $500$.

Figure \ref{fig:neuralembpic} illustrates an example of a 300-dimension fastText embeddings trained over Wikipedia 2017, reduced in 2-dimension using T-SNE algorithm, you can see that word that usually appears in the same context are plotted nearby.

One may ask, how can we get such embeddings ? while a step by step explanation on how to get such embeddings is beyond the scope of this essay, I will only present the available pre-trained word embedding and skip over the detail on how they were obtained.

\section{Embeddings}
Word embeddings have become an essential part of any Deep-Learning-based natural language processing systems. In this work we will review some mechanisms that have been proposed for the calculation of those embedding as we investigate popular toolkit and embeddings matrices that are available online.
\begin{defn}
	An embedding is a mapping from a categorical variable to a vector of continuous numbers.
\end{defn}
People have been using sparse representation (bag-of-word) to encode textural data and as discussed in the previous section, this lead to a dimensionality problem. In the need for reducing the size of such representation and capturing the \textit{core features}, \cite{Bengio03aneural} first introduced the token embedding well known as \textbf{word embeddings}.

To illustrate the idea behind those embeddings, lets look at this example : 

Let us say we are given the following list of courses 

\textbf{\{"C++", "Functional Analysis", "Java", "Swift", "Differential Equation", "Physics", "Python", "Soliton"\}}

Firstly lets try to separate these course, using a line in a way that most similar courses are near each other.
\begin{figure}[H]
	\centering
	\includegraphics[width=0.65\linewidth]{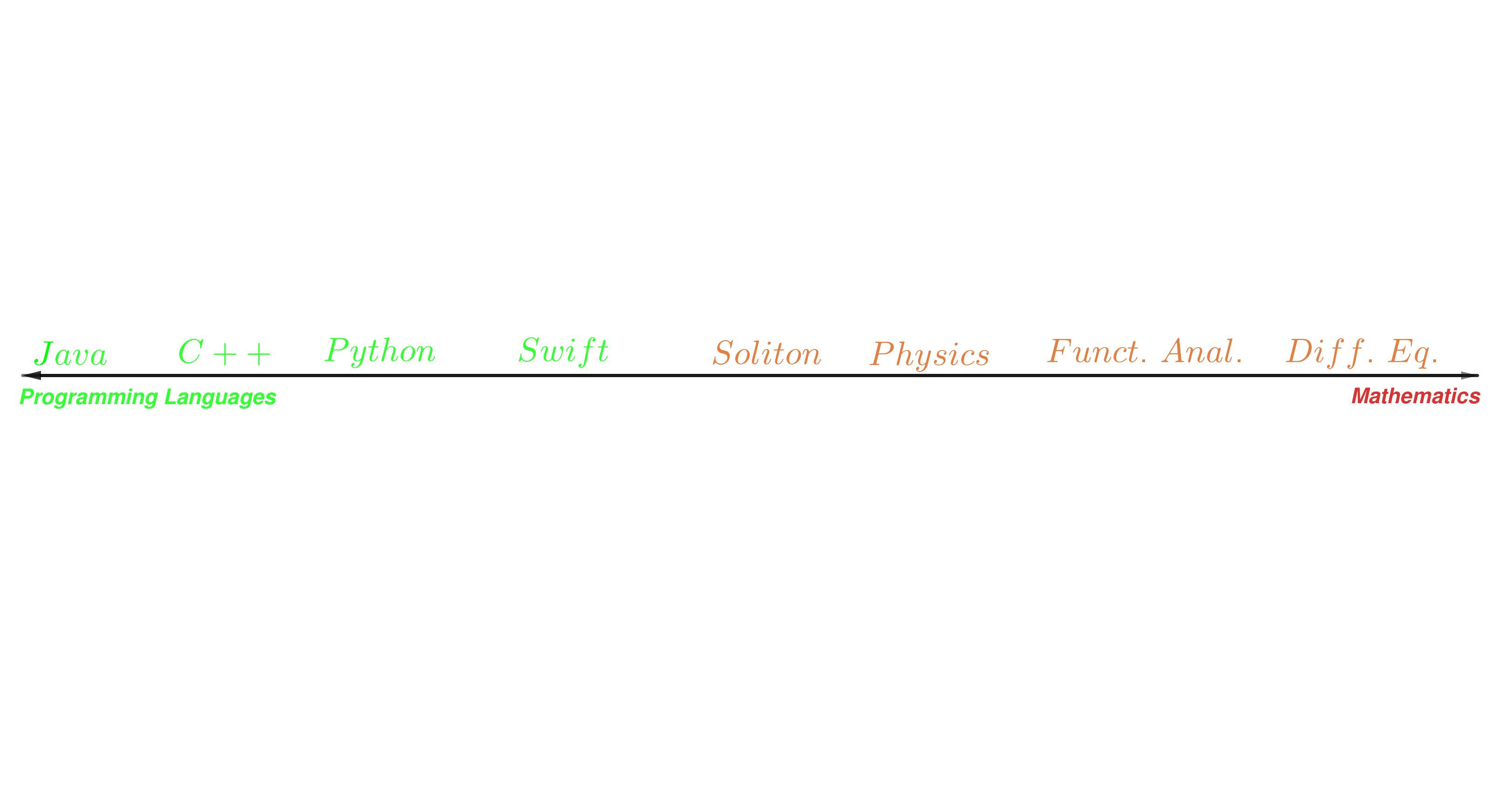}
	\caption{1-D items classification}
	\label{fig:1d-embb}
\end{figure}

As you can see from Figure \ref{fig:1d-embb}, we can kind of put our course in two classes :
\begin{itemize}
	\item Mathematics 
	\item Programming Languages
\end{itemize}

\begin{figure}[H]
	\centering
	\includegraphics[width=0.9\linewidth]{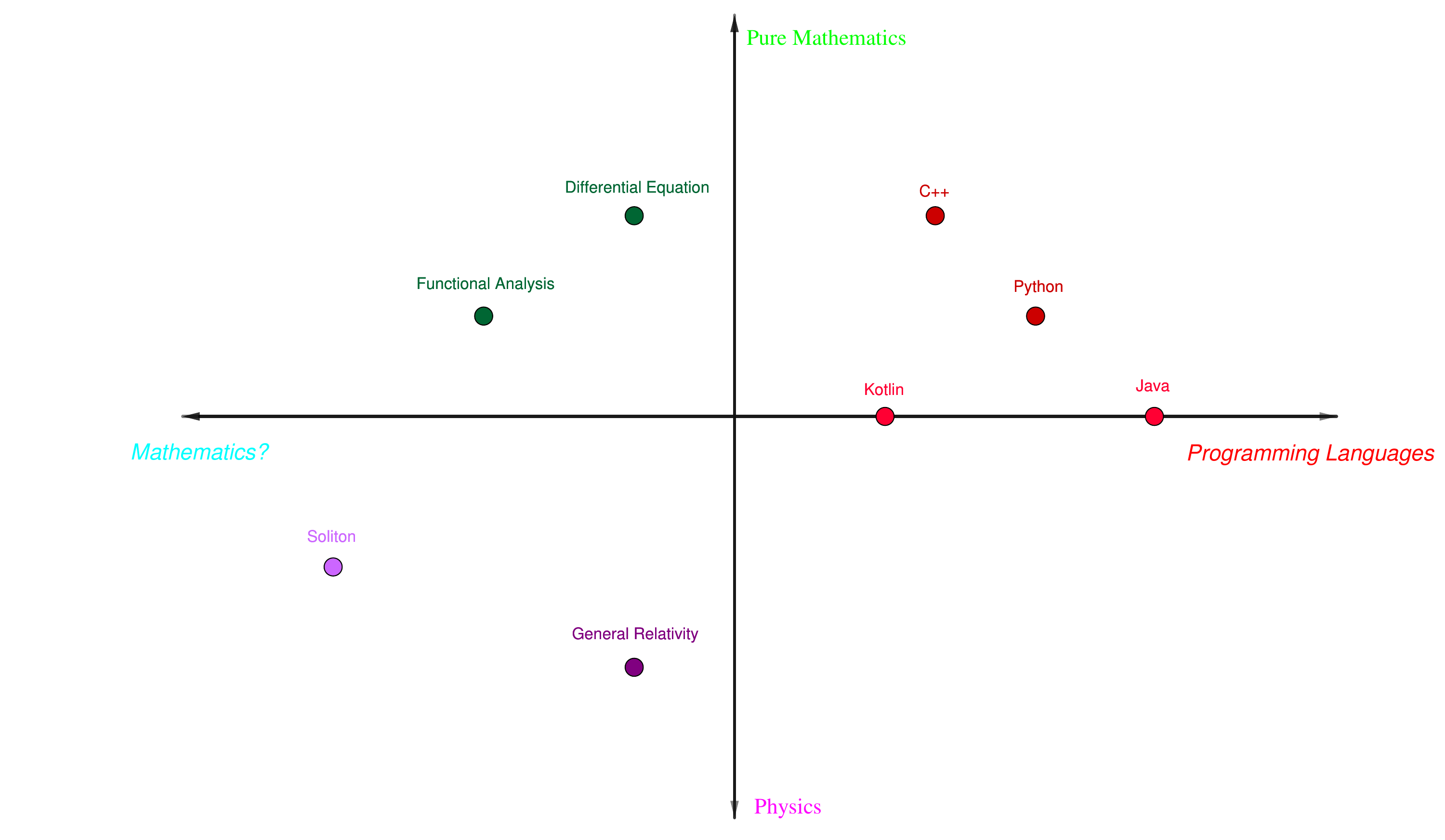}
	\caption{2-D items classification}
	\label{fig:2d-embb}
\end{figure}

Let look closely at our classification problem, one may like to separate pure math courses and pure Physics courses. It's impossible to do such a classification in one line (1-D), the easiest way will be to increase the dimension (2-D). See Figure \ref{fig:2d-embb} for a possible classification.

From the previous analyses and exploration, generally speaking the more we increase a dimension the more we capture a specific feature of our dataset, see Figure \ref{fig:embedding} for a 300-dimension embeddings. This is the general idea in the construction of embeddings.

Next section will presen different type of words representation.

\section{Bag of Words} 
Bag of words(BOW) is a type of representation used in NLP, that represent a text---paragraph, sentence--- as the \textit{bag} of its words, disregarding grammar and even word order but keeping multiplicity.

BOW model is mostly used as a tool for feature generation. After putting every word in it's bag, we calculate various measures to characterize the text.

The next subsection will present the popular ones :

\subsection{Term Frequency (tf)}
this type of BOW aims to calculate the number of times a term appears in the text, this technique is based on the hypothesis that term with high frequency are the most important. 

The term frequency formula is :
\begin{equation}
	tf(word)=\frac{term\ instances}{total\ terms},
	\label{tf_formula}
\end{equation}

For example let consider the two simple text document :
\begin{enumerate}
	\item \textit{Salomon likes to learn new things. Djeff likes that too.} 
	\item \textit{Salomon also likes to learn Karateka.}
\end{enumerate}

\begin{table}[H]
	\centering
	\begin{tabular}{|l|c|c|c|c|c|c|c|c|c|c|c|}
	\hline 
	\textbf{word}& \textit{likes} & \textit{salomon} & \textit{to} & \textit{learn} & \textit{new} & \textit{things} & \textit{djeff} & \textit{that} & \textit{too} & \textit{also} & \textit{Karateka}\\ 
	\hline 
	\textbf{word index} & $1$ & $2$ & $3$ & $4$ & $5$ & $6$ & $7$ & $8$ & $9$ & $10$ & $11$\\ 
	\hline 
	\textbf{BOW\_tf 1} & $0.2$ & $0.1$ & $0.1$ & $0.1$ & $0.1$ & $0.1$ & $0.1$ & $0.1$ & $0.1$ & $0$ & $0$ \\ 
	\hline 
	\textbf{BOW\_tf 2} & $0.17$ & $0.17$ & $0.17$ & $0.17$ & $0$ & $0$ & $0$ & $0$ & $0$& $0.17$ & $0.17$\\ 
	\hline 
\end{tabular}
\caption{Bag of words (tf) representation of sentences "\textit{Salomon likes to learn new things. Djeff likes that too}", "\textit{Salomon also likes to learn Karate}"}
\label{table_bow}
\end{table} 

The following array, represent the bag of word representation of our example, as usually used in pratics  
\begin{align*}
	array([&[0.        , 0.2       , 0.1       , 0.1       , 0.1       ,
	0.1       , 0.1       , 0.1       , 0.1       , 0.1       ,
	0.        , 0.        ],\\
	&[0.        , 0.17, 0.17, 0.17, 0.17,
	0.        , 0.        , 0.        , 0.        , 0.        ,
	0.17, 0.17]])
\end{align*}
The $0$-th index, is always reserved for unknown or rare words.

This kind of representation produces several hard-to-beat performance on task like \textit{email filtering}, but one of the disadvantage of this representation is that we lose any indication of word order, which can be very important for several NLP tasks. 

This representation technique lead quickly to a problem, as common words like \textit{"the"}, \textit{"a"}, \ldots are almost always with highest frequency in texts and thus bias the importance of these words.

To address this problem, \textit{Term Frequency - Inverse Document Frequency} was proposed. 

\subsection{Term Frequency - Inverse Document Frequency (tf-idf)}
This word representation technique is basically the product of the tf(see Equation \ref{tf_formula}) and Inverse document frequency of each word.

The Inverse document frequency of each word is define as :
\begin{equation}
	idf(word) = \log(N/(occurence\ of\ word\ in\ corpus)),
	\label{idf_formula}
\end{equation}
where $N$ is the count of corpus.

As introduced in the previous representation words like \textit{"the"} and \textit{"end"} will usually appear very frequently in every document, the idea with tf-idf is to down weight the importance of these words, by multiplying the term frequency by the inverse of the whole document frequency. But since a corpus may be very large, we usually take the \textit{logarithm} of the inverse document frequency.

So, the \textbf{tf-idf} is basically 
\begin{equation}
	tf-idf(word) =tf(word)\cdot idf(word)
	\label{tf_idf}
\end{equation}
Where $tf(word)$ and $idf(word)$ define as in \ref{tf_formula} and \ref{idf_formula}, respectively .

The tt-idf is often used as a weighting factor is several NLP task, including "sentiment analysis", as will be used in the last Chapter.

Tf-idf requires to load all the text in the document into the memory, to be able to be used effectively and this is sometimes views as a disadvantage of this text representation. 

The following array, represent the tf-idf representation of our previous example,
\begin{align*}
array([&[0., 0., 0., 0., 0., 0.03, 0.03, 0.03, 0.03, 0.03, 0., 0.],\\
&[0., 0., 0., 0., 0., 0.03, 0., 0., 0., 0., 0.05, 0.05],
\end{align*}
again the $0$-th index, is reserved for unknown or rare words.

So far, we have not consider order of words to be relevant in creating words embedding, next section will introduce the approach that brought neural networks in the world of words representation.

\section{Word2Vec}

Word2vec is a model that produces word embeddings. Created by researchers at google \citep{word2vec_Mikolov:2013} in 2013, it is made of two-layer neural networks that are trained to reconstruct linguistic contexts of words. The key behind this unsupervised approach is to produce embeddings in such a way that similar words have similar vectors, this similarity is measure by the idea of \textit{words are similar if they appear in similar contexts}, this is inspired by the well known citation of a British linguist that says "\textit{You shall know a word by the company it keeps}"

The basic idea, is to create embeddings that capture relational aspect of words. For example we will like to have embeddings such that the following qualities hold. 
\begin{align*}
	&\bullet\ "king" -"man" +"woman"="queen"\\
	&\bullet"nature"+"science" = "biology"
\end{align*}

There are two architectural models, Skip-gram and CBOW that are used in the construction of word2vec. The next lines will give a description of the \textit{two} models.

\subsection{Continuous bag-of-word model (CBOW)}
\begin{figure}[H]
	\centering
	\includegraphics[width=0.58\linewidth]{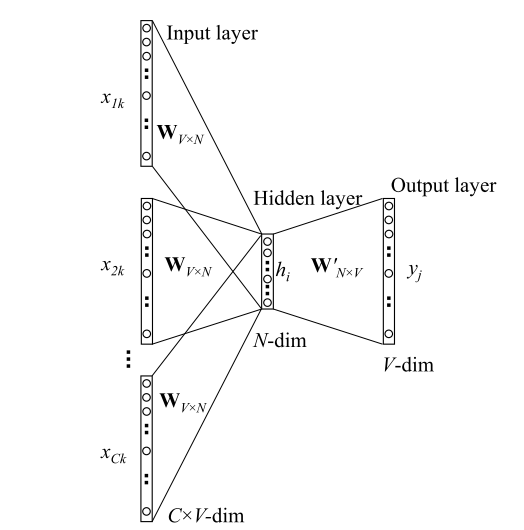}
	\caption{\textbf{CBOW} predict the context given the word}
	\label{fig:cbow}
\end{figure}

Continuous bag-of-word model (CBOW), aims to predict a word known as the \textit{target word} given the \textit{context of that word}, with context defined as words that appear usually together with our target word. Figure \ref{fig:cbow} shows the network under the continuous bag-of-word model. In our sitting, the vocabulary size is $V$, and the hidden layer size is $N$, all nodes are fully connected to each others. The input is a one-hot encoded vector i.e for a given input context word, only one out of $V$ units is $1$.

The weights between the input layer and the hidden layer is a $V\times N$ matrix $\mathbf{W}$. Each row of $\mathbf{W}$ is the $N$-dimension vector representation $\mathbf{v}_{w}$ of the associated word of the input layer. 

\begin{align*}
	\mathbf{h}&=\frac{1}{C}\mathbf{W}^T (\mathbf{x}_1+\mathbf{x}_2 +\cdots+\mathbf{x}_C)\\
	 		  &=\frac{1}{C} (\mathbf{V}_{W_{1}}+\mathbf{V}_{W_{2}} +\cdots+\mathbf{V}_{W_{c}})^T
\end{align*} 
where $C$ is the number of words in the context, $w_1,\cdots,w_C$ are the words in the context, and $\mathbf{v}_w$ is the input vector of a word $w$.

From the hidden layer to the output layer, we have an other weights matrix denoted $\mathbf{W}^\prime$\footnote{$\mathbf{W}^\prime$ here mean an other matrix, not the derivative or the the transpose of $\mathbf{W}$}, which is an $N\times V$ matrix. Using these weights, we obtain a score $u_j$ for each word in the vocabulary,
\begin{equation}
	u_j = \mathbf{v}^{\prime\quad T}_{w_j} \mathbf{h},
\end{equation}
with $\mathbf{v}^\prime_{w_j}$ the $j$-th column of the matrix $\mathbf{W}^\prime$. 

Using softmax, a log-linear classification model, 

\begin{equation}
y_{j} = p\left(w_{j} | w_{I}\right)=\frac{\exp \left(u_{j}\right)}{\sum_{j^{\prime}=1}^{V} \exp \left(u_{j^{\prime}}\right)},
\end{equation}
with $y_j$ the output of the $j$-th unit in the output layer.

\subsection{skip-gram}
\begin{figure}[H]
	\centering
	\includegraphics[width=0.5\linewidth]{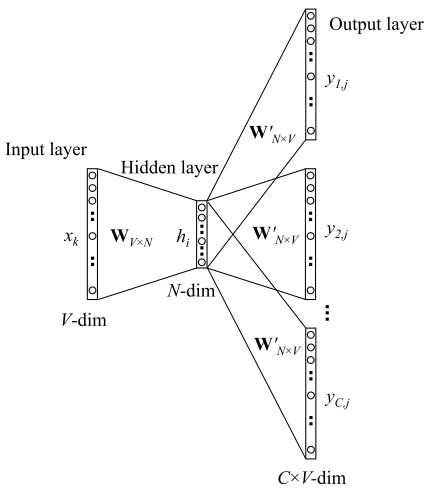}
	\caption{\textbf{Skipgram}, predict a word given the context}
	\label{fig:skip-gram}
\end{figure}

The \textbf{skip-gram} architecture, is basically the inverse of the CBOW as can be seen in Figure \ref{fig:skip-gram}, here the input is a word---\textit{target word}--- $w$ and output the words that often surround that target word. For example let us consider the sentence :"\textit{I like the machine learning}", lets assume the input to be "\textit{the}" and consider a widow size of $2$, the output is "\textit{I}", "\textit{like}", "\textit{machine}" and "\textit{learning}".

The input vector on the input layer will still be denoted by $\mathbf{v}_{wI}$, because the input vector is one-hot encoded, $\mathbf{h}$ is simply copying the row of $\mathbf{W}$, associated with $w_I$.
\begin{equation}
	\mathbf{h}=\mathbf{W}_{(k, \cdot)}^{T} :=\mathbf{v}_{w_{I}}^{T}
\end{equation}

On the output side, the output layer panels share the same weights $\mathbf{v}^{\prime}_w$,
\begin{equation*}
	u_{c, j}=u_{j}=\mathbf{v}_{w_{j}}^{\prime\quad T}  \mathbf{h}, \text {  for  } c=1,2, \ldots, C
\end{equation*}
where $\mathbf{v}^{\prime}_{w_j}$ is the output vector of the $j$-th word in the vocabulary $w_j$ and $u_{c,j}$ is the network input of the $j$-th unit on the $c$-th panel of the output layer.

Finally,
\begin{equation}
	y_{c, j} = p\left(w_{c, j}=w_{O, c} | w_{I}\right)=\frac{\exp \left(u_{c, j}\right)}{\sum_{j^{\prime}=1}^{V} \exp \left(u_{j^{\prime}}\right)}
\end{equation}
where $w_{c,j}$ is the $j$-th word on the $c-$th panel of the output layer, $w_{O,c}$ is the output of the $j$-th neuron on the $c$-th panel of the output layer. 

\section{GloVe}
Global Vectors (GloVe) is an unsupervised learning algorithm as word2vec but was proposed by Stanford as an improved version of word2vec.

Training is performed on aggregated global word-word co-occurrence statistics from a corpus \citep{glove_pennington2014}. 

GloVe learn word representations via matrix factorization, it minimizes the difference between the dot product of the embeddings of a word $w_i$ and its context word $c_t$ and the logarithm of their number of co-occurences within a certain window size :
\begin{equation}
\mathcal{L}_{\text { Glove }}=\sum_{i, j=1}^{|V|} f\left(\mathbf{C}_{i j}\right)\left(\mathbf{x}_{i}^{\top} \mathbf{x}^{\prime}_{j}+b_{i}+b^{\prime}_{j}-\log \mathbf{c}_{i j}\right)^{2}
\end{equation}
where $f(\cdot)$ is a weighting function that assigns relatively lower weight to rare ans frequent co-occurrences, $c_{ij}$ represent the number of times word $w_i$ appears with context word $w_j$, and $b_i$ and $b^\prime$ are biases corresponding to word $w_i$ and its context word $w_j$ 
\section{fastText}
Created by Facebook \citep{fasttext_bojanowski2017} to overcame the limitation of word2vec and glove.  word2vec and glove consider each word as single unit and ignore the morphological structure of the word the are unable to generate word embedding for the unseen or out of vocabulary word during the training, this ability of fastText is due to the fact that it considers $n$-gram of characters and the global representation of a token can be seen as the sum of $n$-gram characters. For example, with $n = 3$ the word "\textit{where}" would be represented by the character n-grams: <\textit{wh, whe, her, ere, re}>

fastText have been demonstrated to produce better results than the previous two, we will in the next Chapter report an experimentation using this technique

\section{Our Model Architecture}
Training an embedding only makes sense when there is a lot of training data available, in this work we will use pre-trained embeddings that are trained on a huge amount of data will be used. 

The input $x$ in NLP task is usually words, part of speech tags or other linguistic information. In our case the input is a paragraph of words, that represent someone's review on a specific movie. 
\begin{figure}
	\centering
	\includegraphics[width=0.7\linewidth]{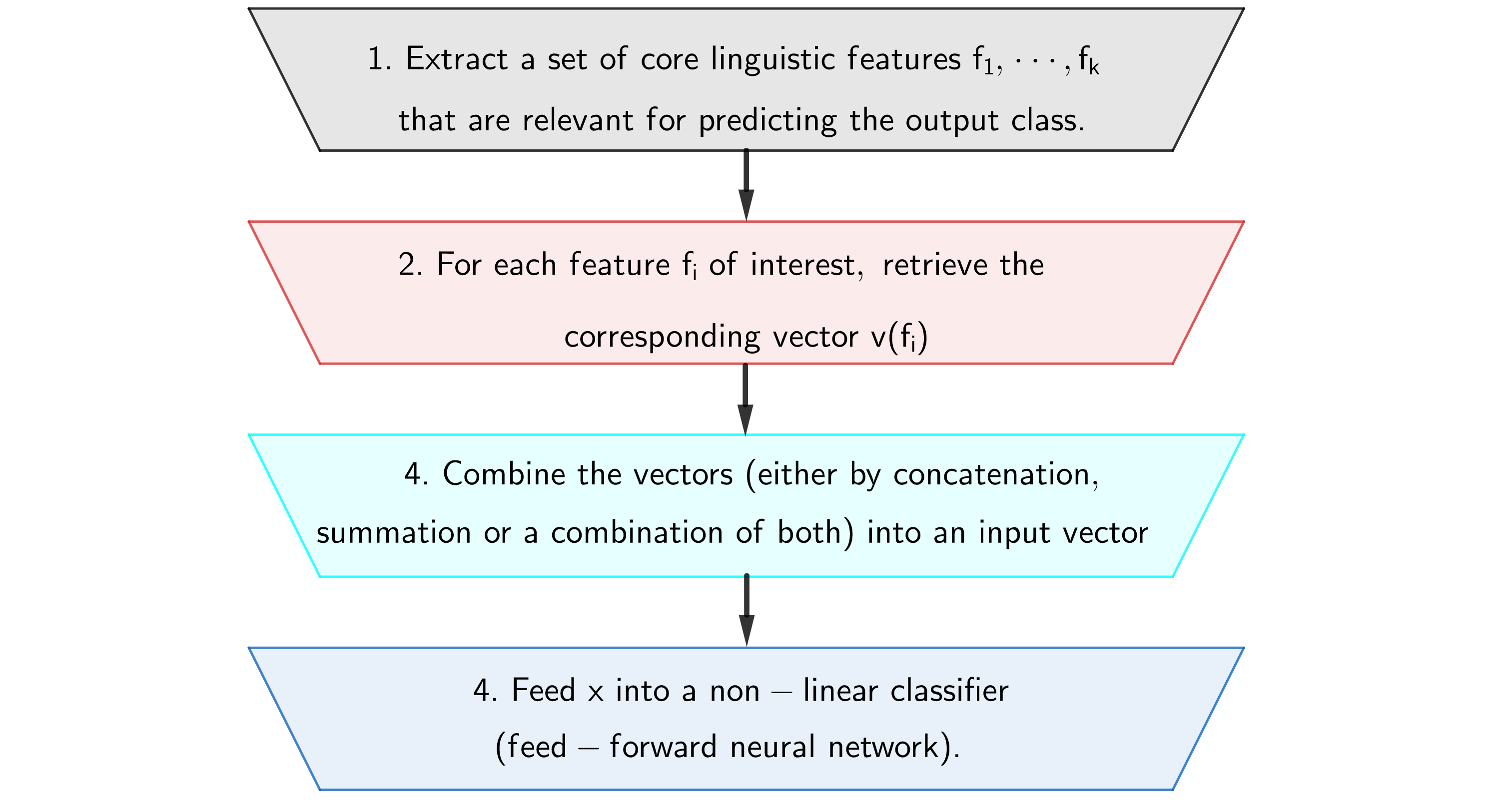}
	\caption{General structure for an NLP classification system based on a feed-forward neural network}
	\label{fig:nlpclassification}
\end{figure}

Figure \ref{fig:nlpclassification} describes the general structure for an NLP classification system based on feed-forward neural network \cite{Goldberg:2016:PNN:3176748.3176757}, at step 2, we will use the pre-trained word embeddings to get the vector representation of features.

Let $c(\cdot)$ be a function from the core features to the input vector $x$ in our neural network.

Let $x$ be of shape $(N, n)$, with $N$ the number of examples and $n$ the vocabulary size.

Lets consider as well $f_{ij}$ be the feature (token) $i$ in the example $j$, with $i:\{0,1, \cdots,n \}$ and $j:\{0,1,\cdots,N-1 \}$.
\begin{align*}
	x_{j} &= c(f_{0j}, f_{1j}, f_{nj})\\
		&= WCBOW\left(f_{0j}, f_{1j},f_{nj}\right)\\
		&=\frac{1}{\sum_{i=1}^{n} a_{ij}}\sum_{i=1}^{n}a_{ij} v\left(f_{ij} \right)\\
\end{align*}
With $WCBOW$ the weighted continuous bar of words $(CBOW)$, is a very similar to the traditional bag of words representation in which different vectors receive different weights, in our case the weight is the word's TF\_IDF (See Equation \ref{tf_idf}).

Given $f_i$ the value of $v(f_i)$ depend to the type of pre-trained word embeddings one may choose to use. In this essay we will do our exploration using fasttext word embeddings. 

\subsection{Evaluation metric} NlP systems are evaluated with usually with regard to their performance on the test set of the specific task. 

\textbf{Accuracy}
Accuracy is one metric for evaluating classification models.   

\begin{equation}
A c c=\frac{T P+T N}{T P+T N+F P+F N}
\end{equation}
where $TP$ = True Positives, $TN$ = True Negatives, $FP$ = False Positives, and $FN$ = False Negatives. 

We can summarize the above measure in a $2\times2$ matrix called \textit{confusion matrix}, that can hep in getting insight in how our model performs.

\textbf{Receiver Operating Curve (ROC) Curve and Area Under the Curve (AUC)}
A ROC is a graph that shows the performance of a classification model at all classification thresholds.

For efficiency reason, AUC measures the entire two-dimensional area under the ROC curve instead of evaluating the model many times with different classification thresholds. 

One easy way of interpreting the AUC is the probability that the model rank a random positive example more highly than a random negative example.

We will use these evaluation accuracy in the next chapter, to measure the performance of our classifier.

The next chapter will now review and do some sentiment analysis experiment using the well known \textit{Internet Movie Database}(IMDb) data set.

%% file: chapter4.tex
\chapter{Results and Discussion}\label{chap4}
In this chapter we present the results, analyze and discuss an investigation on a sentiment analysis task using a simple neural networks model.


%
%
%
%

\section{Data preprocessing}
Data preprocessing is a crucial step when building a machine learning model. In the traditional programming debugging a program mean analyzing the code in order to find what went wrong, but contrary a machine learning debugging consist of analyzing the data presented to the model. As good as a model can be, bad preprocessing can produce unexpected output.

\subsection{Gather Data}
We will investigate on  IMDb(Internet Movie Database) data set. We will use a specific version of this dataset\footnote{http://ai.stanford.edu/\~amaas/data/sentiment/} as made available by \cite{dataset:2011:ACL-HLT2011}, this dataset consists of customer reviews about a movie, those reviews are manually labeled by positivity from 1 to 10 and polarity as \textbf{positive} or \textbf{negative}, for this work we are interested only in the question : \textbf{How can we get the polarity---sentiment---of a customer given only is review ?}


\subsection{Explore Data}
We observed that reviews contains noises due to the presence of incomplete sentences or words, misspell words and out of dictionary words. Thus we will clean all the reviews by removing all the html tags and punctuation---normalization.

\begin{figure}[H]
	\centering
	\includegraphics[width=0.9\linewidth]{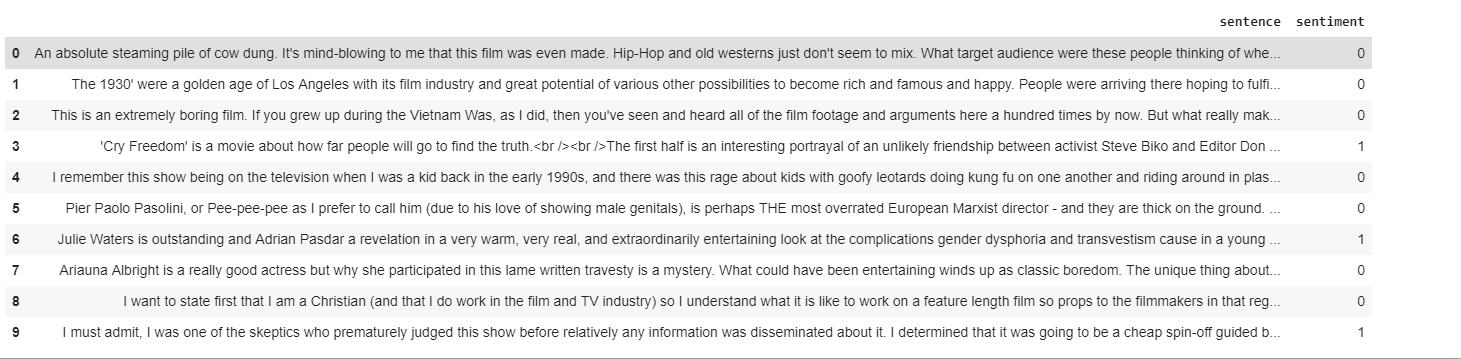}
	\caption{First 10 items in the training set}
	\label{fig:view_train}
	\includegraphics[width=0.9\linewidth]{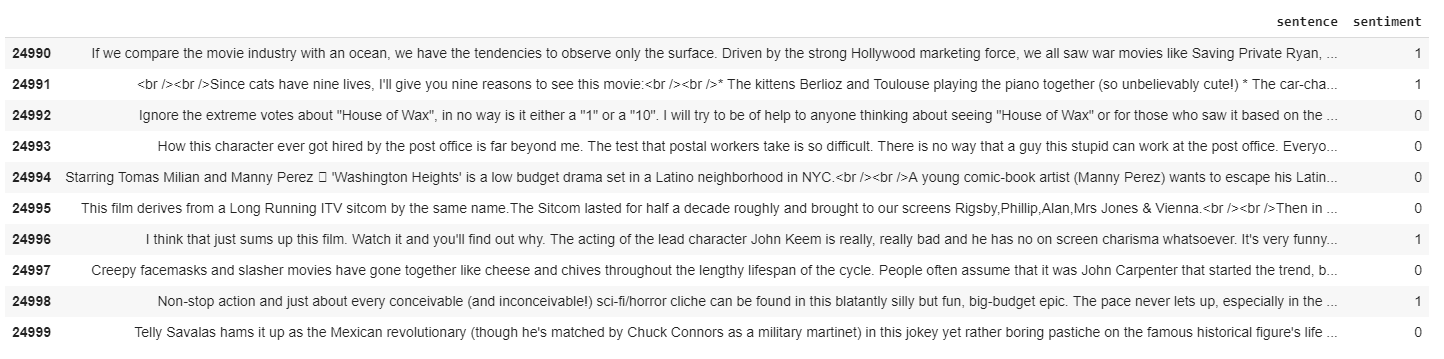}
	\caption{Last 10 items in the testing set}
	\label{fig:viewtest}
\end{figure}

We have a set of $25,000$ highly polar movie reviews for training, and 25,000 for testing, Figure \ref{fig:view_train} and \ref{fig:viewtest} gives a sample of how our data looks like in the training set and in the testing set, respectively.  
Additional, our data is well balanced, in both the training and testing set, $12500$ are positive and the remaining one are negative.  


%
%
%
%

\section{Experimentation and Results}
The result obtain in this section are obtains based on a neural network with $3$ \textit{hidden layer} that use $relu$ activation function. Every layer have $512$ \textit{units}, we are using \textit{binary\_crossentropy} as our loss function and \textit{adamax} as our optimizer with a learning rate of $0.003$. 

After pre-processing, our data contains $88582$ unique tokens, but due to storage and computational limitation we first did our investigation using only $20000$ top words, of course more increasing this value, will eventually affect the accuracy of the model as we will experiment later in this Chapter. 

\begin{figure}
	\centering
	\includegraphics[width=0.44\linewidth]{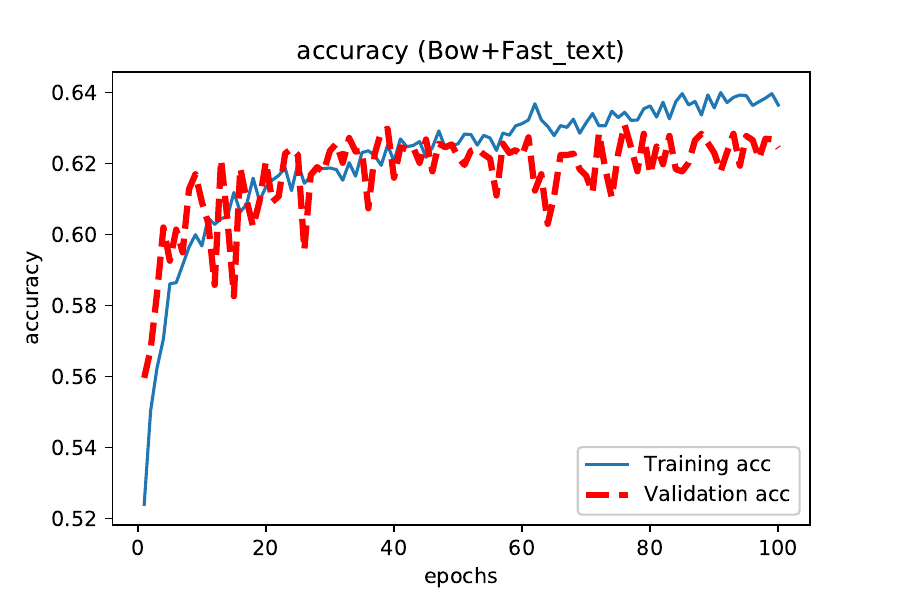}\includegraphics[width=0.44\linewidth]{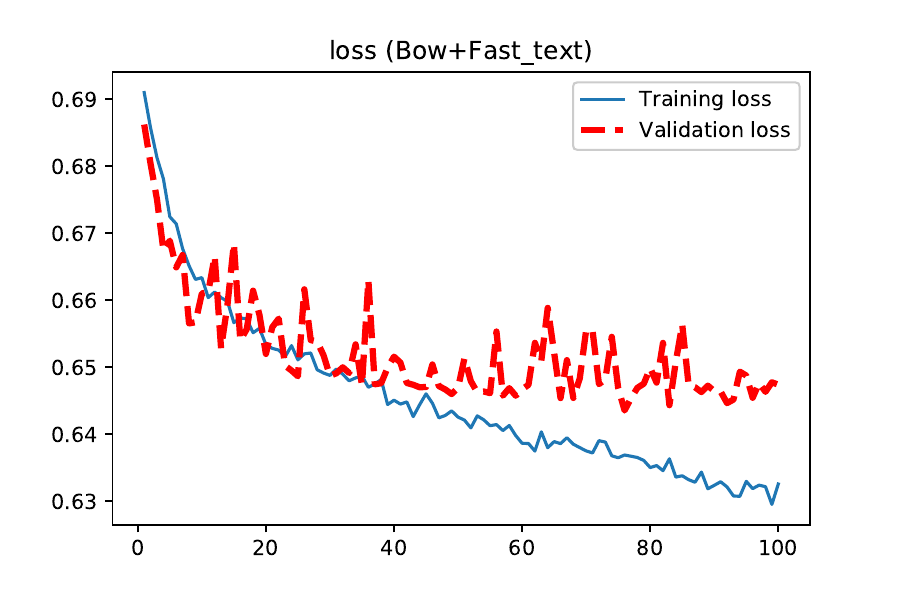}\\
	\includegraphics[width=0.44\linewidth]{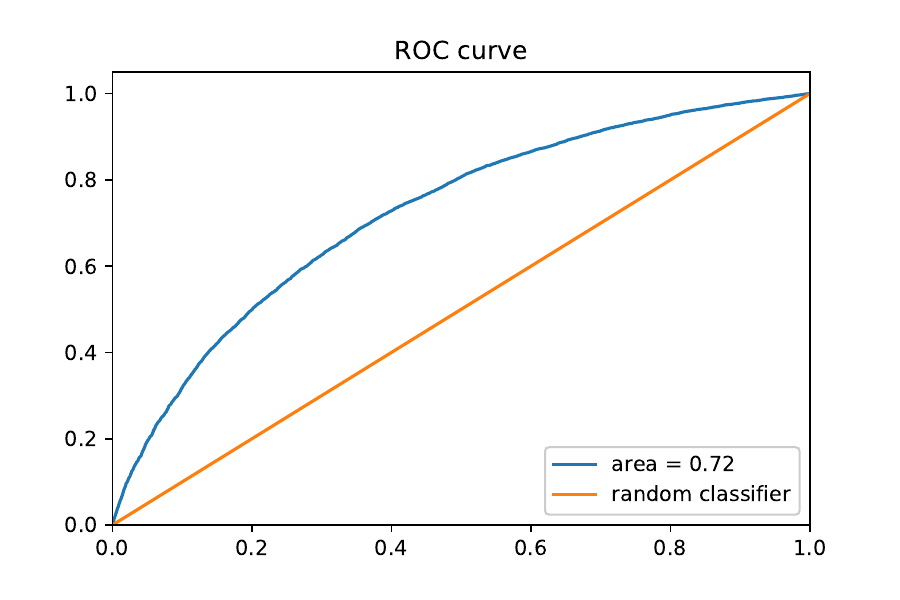}\includegraphics[width=0.44\linewidth]{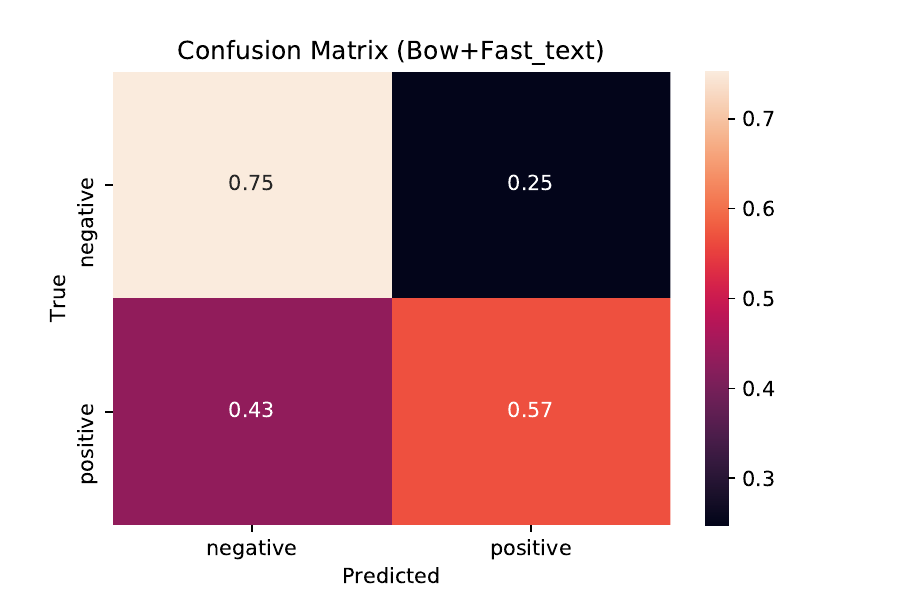}
	\caption{accuracy vs epochs, accuracy vs epochs, Roc Curve, and Confusion Matrix of IMDb using WCBOW+fastText embeddings representation}
	\label{fig:plt_tf}
\end{figure}

Firstly, we started, by using tf-idf---a modified version as implemented in keras pre-processing package---representation as define in Equation \ref{tf_idf} and run our neural network as described above and obtained the accuracy of $79.9\%$ on unseen data, and Figure \ref{fig:plt_tf} reports the learning curve tendency and the confusion matrix plots. 
\begin{figure}
	\centering
	\includegraphics[width=0.44\linewidth]{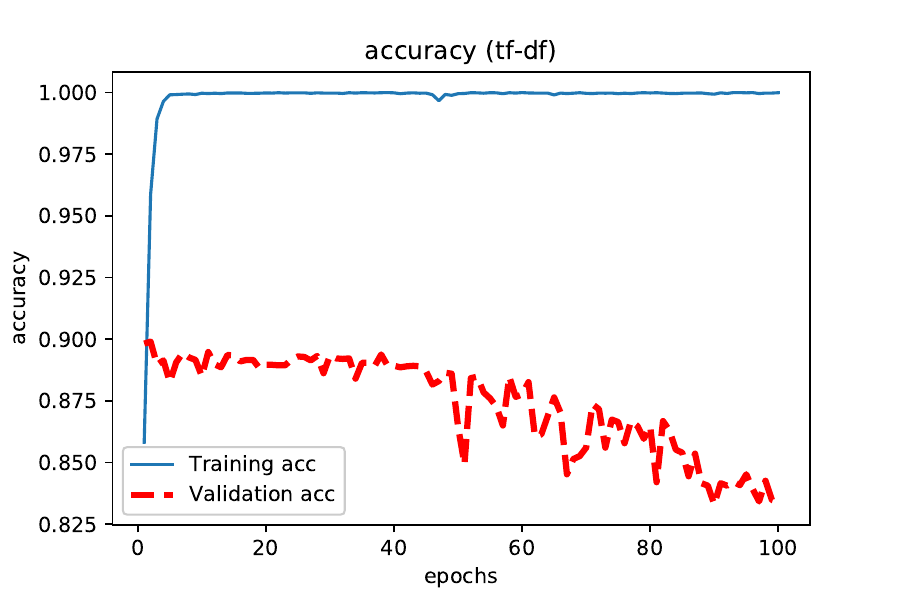}\includegraphics[width=0.44\linewidth]{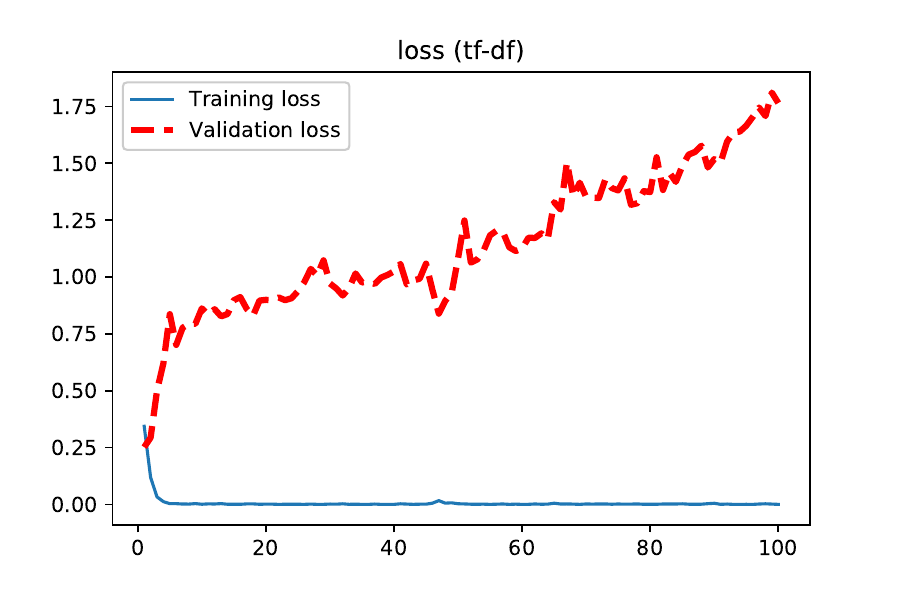}\\
	\includegraphics[width=0.44\linewidth]{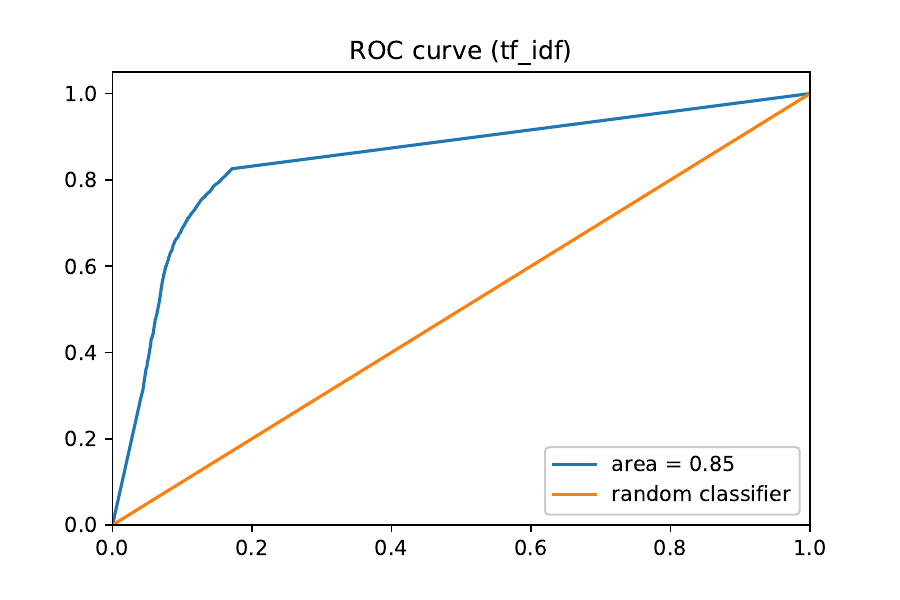}\includegraphics[width=0.44\linewidth]{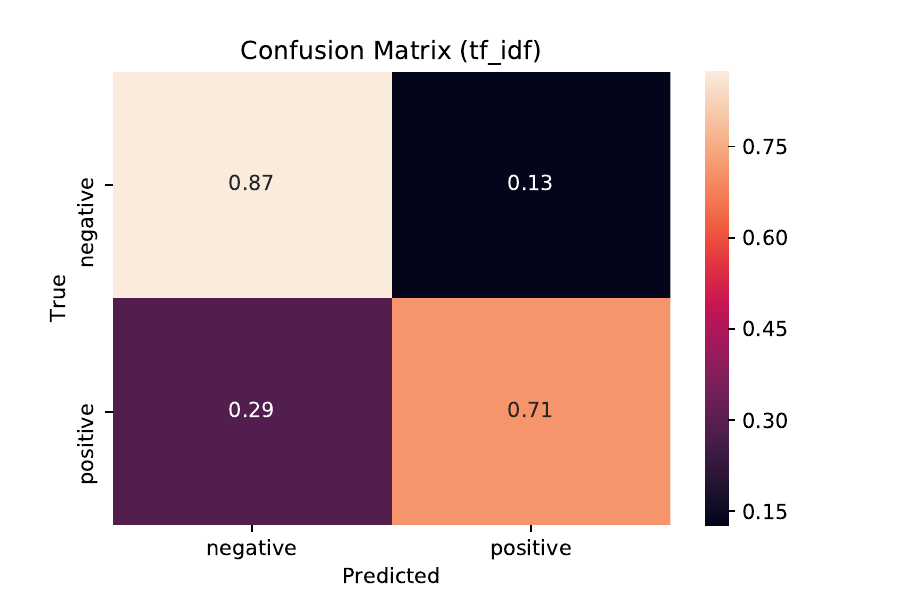}
	\caption{accuracy vs epochs, loss vs epochs, ROC curve, and Confusion Matrix plots obtains using BOW(tf-idf) representation}
	\label{fig:plot_emb}
\end{figure}

Secondly, we used our architecture described in the Chapter 3, to transform the word embeddings of individual word to sentence embeddings of the entire review. 

Out of the $20000$ top words considered, $1883$ were out-of-vocabulary(OVV)---no embeddings found in the fastText pre-trained word embeddings---we at first ignore those OVV words and obtain an accuracy of $66.35\%$ and Figure \ref{fig:plot_emb} report the learning curve tendency and the confusion matrix plots.  
 
Lastly, we used a pre-trained word embeddings module\footnote{https://tfhub.dev/google/Wiki-words-250-with-normalization/1}, create by Google that is freely available on TensorFlow Hub website.TensorFlow hub is a library for the publication, discovery, and consumption of reusable parts of machine learning models, which will do the pre-processing and map all the out-of-vocabulary tokens into one bucket that is initialized with zeros.

This TensorFlow module in based on skipgram version of word2vec trained on English Wilipedia corpus. 

\begin{figure}[H]
	\centering
	\includegraphics[width=0.44\linewidth]{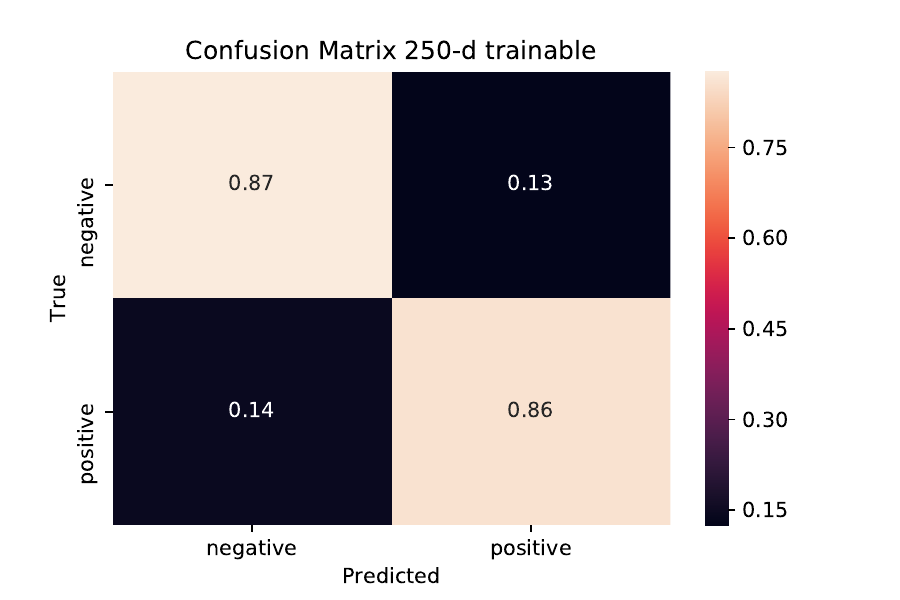}\includegraphics[width=0.44\linewidth]{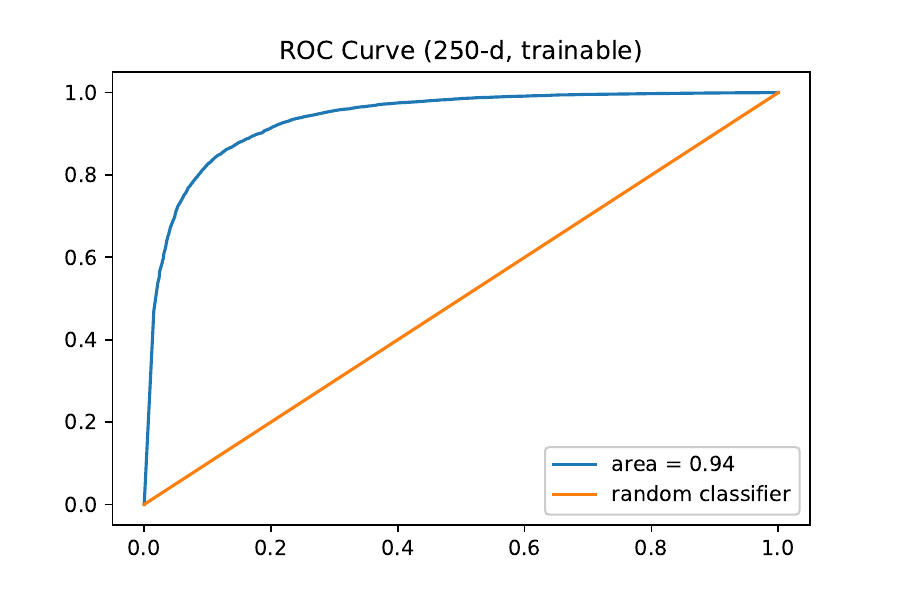}
	\caption{Confusion Matrix and Roc Curve, obtains using the TensorFlow hub module}
	\label{fig:confusion-final}
\end{figure}

This module has allowed us to include more consider more words and we have obtain an accuracy of $86.5\%$ on the testing set and the confusion matrix can be seen in Figure \ref{fig:confusion-final}

\section{Discussion}
In the previous section we have seen the influence of using word embeddings .

The BOW representation, are known to be simple and efficient in topic-based text classification. But as noticed in the result section it is not suitable for sentiment classification because it breaks the syntactic structures, and ignore word order, as well as semantic information. 

Our results based on BOW representation shows that even though at first the accuracy obtained using tf-idf representation look high but the Figure \ref{fig:plt_tf} reveal that there is a problem in our model, the model tend to do very well on the training but not on the validation data---overfitting.

Due to the limitation of BOW, we proposed also to use fastText embeddings and we noticed that even though the accuracy is more lower than the simple BOW, it is more precise, in classifying reviews. (See Figure \ref{fig:confusion-final}).

We can see that encoding sequence of words into weighted sum of the corresponding word embeddings has helped us to capture composition and dependencies of our model features. 

By using the TensorFlow hub module, we could consider more words and handle properly the out-of-vocabulary words. We observed that pre-trained word embeddings gives amazing result and the miss classified example are much more lower than in previous experimentation. (See Figure \ref{fig:confusion-final}) 

The reason of the high accuracy observed in the last implementation is also due to the fact that the embeddings layer is jointly trained with the other layer of our model, this has improved considerably the generalization power of our model. 

We are going in the last part to conclude our work and introduce further works.

%% file: chapter5.tex
\chapter{Conclusion and Further Work}\label{chap5}

\section{Conclusion}
The aim of this essay was to investigate proprieties and the use of word embeddings in Natural Language Processing.
We presented various techniques of word representations and gave a brief overview on methods and techniques used in their construction.
Representation learned using neural networks has proved to outperform classical embedding methods such as approaches based on bag-of-words. One aspect of distinction is that in word2vec for instance, similar word may have similar vector representation, and this ability allows the model to gain more insight of input features.

Our experimentation has demonstrated that pre-trained word embeddings are able to supplement training data and it increases the generalization power of a simple neural networks composed of simple dense layers. The python code used for our investigation can freely be found \href{https://github.com/Kabongosalomon/Word-Embedding-Investigation/blob/master/An\%20empirical\%20investigation\%20into\%20the\%20properties\%20of\%20standard\%20word\%20embeddings.ipynb}{here} (this code suppose the use of the google colaboratory notebook).

Throughout our investigation we have noticed that many big companies only provide ready to use modules for natural language processing tasks, this has been a great challenge in NLP for young researchers, I would like to add my voice to many others, to ask for promotion of the \textit{open science} initiative.

In conclusion, we experimentally demonstrate that the representation learned using neural networks are able to significantly improve results on the sentiment analysis task.

\section{Further Work}

For future work, we would like to :
\begin{itemize}
	\item use cross-lingual word embedding models, study its propriety for application in machine translation. 
	\item use more advanced neural network layer as $1$-dimension convolution neural network or recurrent neural networks jointly with generalization power that embeddings brings to experiment on several NLP tasks. 
	\item investigate on the possibility of using doc2vec in the creation of embeddings 
	\item propose a module on the TensorFlow hub website that can support Congolese official languages.
\end{itemize}


%% file: acknowledgement.tex
\chapter*{Acknowledgements}
First, I would like to thank the lord El-Shadda\"i the one I serve for giving me this opportunity to complete this degree.

Secondly, I would like to thanks my supervisor Professor Etienne Barnard for been there for me at every step throughout the completion of this project. Thanks to my tutor Jordan Masakuna for reading and commenting on the improvement of my work. Thanks to all AIMS staff and administration for everything. I would also like to thank my fellow AIMS students for their support. Thanks to Irene Kyomugisha for taking to read and give comment on this work.

Thirdly, I would like to give a big thank to Professor Neil Turok, for everything, your vision for Africa has impacted me. I would like to thank also my academic director J.W. Sanders for his support, through talks and presentations. I've really learn a lot from your sir. 

Lastly, special thanks to my biological parents, Kabenamualu Kabongo Frederic and Ngalula wa Lukusa Beatrice, and all my family, your patience and love helped me to get this far. To Abigail thank you for your presence in my life, I love you.

%